\documentclass[sigconf]{acmart}
\usepackage{subfigure}
\usepackage{multirow}
\usepackage{makecell}
\usepackage[normalem]{ulem}
\useunder{\uline}{\ul}{}
\AtBeginDocument{%
  }
\setcopyright{acmcopyright}
\copyrightyear{2018}
\acmYear{2018}
\acmDOI{10.1145/1122445.1122456}

\acmConference[Conference acronym 'XX]{Make sure to enter the correct
  conference title from your rights confirmation emai}{June 03--05,
  2018}{Woodstock, NY}
\acmPrice{15.00}
\acmISBN{978-1-4503-XXXX-X/18/06}

\newboolean{showComments}
\setboolean{showComments}{false}
\newcounter{noteYHctr} 
\ifthenelse{\boolean{showComments}}{
    \newcommand{\YHd}[1]{ \textbf{\textcolor{orange}{{{Sheryl: \#\arabic{noteYHctr}: }}#1}} \addtocounter{noteYHctr}{1} }
}{ \newcommand{\YHd}[1]{{}} 
}

\ifthenelse{\boolean{showComments}}{
    \newcommand{\YH}[1]{ \textbf{\textcolor{orange}{{{Sheryl: \#\arabic{noteYHctr}: }}#1}} \addtocounter{noteYHctr}{1} }
}{
 \newcommand{\YH}[1]{}
}

\newcommand{\yh}[1]{{\textcolor{black}{{{}}#1}}{}}

\definecolor{purple}{RGB}{178,55,250} 

\newcounter{noteGQctr} 
\ifthenelse{\boolean{showComments}}{\newcommand{\GQ}[1]{ \textbf{\textcolor{brown}{{{Quan: \#\arabic{noteGQctr}: }}#1}} \addtocounter{noteGQctr}{1} } }{
 \newcommand{\GQ}[1]{}
}
\ifthenelse{\boolean{showComments}}{
    \newcommand{\GQd}[1]{ \textbf{\textcolor{brown}{{{Quan: \#\arabic{noteGQctr}: }}#1}} \addtocounter{noteGQctr}{1}
 }
 }{
 \newcommand{\GQd}[1]{}
}

\newcommand{\gq}[1]{{\textcolor{black}{{{}}#1}}{}}

\ifthenelse{\boolean{showComments}}{
\newcommand{\JC}[1]{ \textbf{\textcolor{green}{{{Junchi: \#\arabic{noteJCctr}: }}#1}} \addtocounter{noteJCctr}{1} }}{
 \newcommand{\JC}[1]{}
}
\ifthenelse{\boolean{showComments}}{
    \newcommand{\JCd}[1]{ \textbf{\textcolor{green}{{{Quan: \#\arabic{noteJCctr}: }}#1}} \addtocounter{noteJCctr}{1}
 }
 }{
 \newcommand{\JCd}[1]{}
}

\ifthenelse{\boolean{showComments}}{
\newcommand{\XH}[1]{ \textbf{\textcolor{red}{{{Xuhong: \#\arabic{noteJCctr}: }}#1}} \addtocounter{noteXHctr}{1} }}{
 \newcommand{\XH}[1]{}
}
\ifthenelse{\boolean{showComments}}{
    \newcommand{\XHd}[1]{ \textbf{\textcolor{ref}{{{Xuhong: \#\arabic{noteXHctr}: }}#1}} \addtocounter{noteXHctr}{1}
 }
 }{
 \newcommand{\XHd}[1]{}
}

\newcommand{\xh}[1]{{\textcolor{black}{{{}}#1}}{}}

\begin{document}
\title{CEP3: Community Event Prediction with Neural Point Process on Graph}


\author{Xuhong Wang}
\authornote{Both authors contributed equally to this research.}
\authornote{This work has been completed during their internship in Amazon.}
\email{wang\_xuhong@sjtu.edu.cn}
\affiliation{%
  \institution{Shanghai Jiao Tong University}
  \city{Shanghai}
  \country{China}}

\author{Sirui Chen}
\authornotemark[1]
\authornotemark[2]
\email{ericcsr@connect.hku.hk}
\affiliation{%
  \institution{The University of Hong Kong}
  \city{Hong Kong}
  \country{China}
  \postcode{43017-6221}
}

\author{Yixuan He}
\authornotemark[2]
\email{yixuan.he@stats.ox.ac.uk}
\affiliation{%
  \institution{University of Oxford}
  \city{}
  \country{United Kingdom}
}

\author{Minjie Wang}
\email{minjiw@amazon.com}
\affiliation{%
  \institution{Amazon}
  \city{Shanghai}
  \country{China}
}

\author{Quan Gan}
\authornote{Corresponding author}
\email{quagan@amazon.com}
\affiliation{%
  \institution{Amazon}
  \city{Shanghai}
  \country{China}
}

\author{Yupu Yang}
\email{ypyang@sjtu.edu.cn}
\affiliation{%
  \institution{Shanghai Jiao Tong University}
  \city{Shanghai}
  \country{China}}

\author{Junchi Yan}
\email{yanjunchi@sjtu.edu.cn}
\affiliation{%
  \institution{Shanghai Jiao Tong University}
  \city{Shanghai}
  \country{China}}

\renewcommand{\shortauthors}{Xuhong and Sirui, et al.}


\begin{abstract}

Many real world applications can be formulated as event forecasting on Continuous Time Dynamic Graphs (CTDGs) where the occurrence of a timed event between two entities is represented as an edge along with its occurrence timestamp in the graphs.However, most previous works approach the problem in compromised settings, either formulating it as a link prediction task on the graph given the event time or a time prediction problem given which event will happen next. In this paper, we propose a novel model combining Graph Neural Networks and Marked Temporal Point Process (MTPP) that jointly forecasts multiple link events and their timestamps on communities over a CTDG. Moreover, to scale our model to large graphs, we factorize the jointly event prediction problem into three easier conditional probability modeling problems.To evaluate the effectiveness of our model and the rationale behind such a decomposition, we establish a set of benchmarks and evaluation metrics for this event forecasting task. Our experiments demonstrate the superior performance of our model in terms of both model accuracy and training efficiency. 
\end{abstract}



\maketitle 
\section{Introduction}



\xh{Modeling and learning from dynamic interactions of entities has become an important topic and inspired a great interest in wide application fields.
\gq{In particular, \xh{studying the evolution of community social events can enable preemptive intervention for the pandemic (e.g., COVID-19) spreading~\cite{healthcare2}. Monitoring and forecasting the traffic congestion~\cite{dcrnn} spreading can help the police prevent congestion from spreading outside the community. The community needs more attention and help if there is a sudden change in the state of the economic~\cite{economics} or political leanings~\cite{socialevolve}. }}
In some cases, \yh{some entities, such as those } with dense connections\yh{, or with similar characteristics,} may form certain communities\yh{, and communities could also be defined by users based on their criteria.} 
\GQd{Do we only care about those with dense connections?  I will probably just remove this sentence.}\YHd{perhaps any kind of community of interest, but the Louvain algorithm used in the experiments consider dense connections} \GQd{I would say Louvain is just experimental detail; probably doesn't matter in introduction.}
\YHd{updated, could check}
\yh{In reality, people may only be interested in }
a specific community of entities’ interactions in some practical applications, such as community behavior modeling~\cite{CoNN}, dynamic community discovery~\cite{DBLP:journals/csur/RossettiC18} and community outliers detection~\cite{DBLP:conf/kdd/GaoLFWSH10}.}

\begin{figure}[tb!]
\centering
  \includegraphics[width=\linewidth]{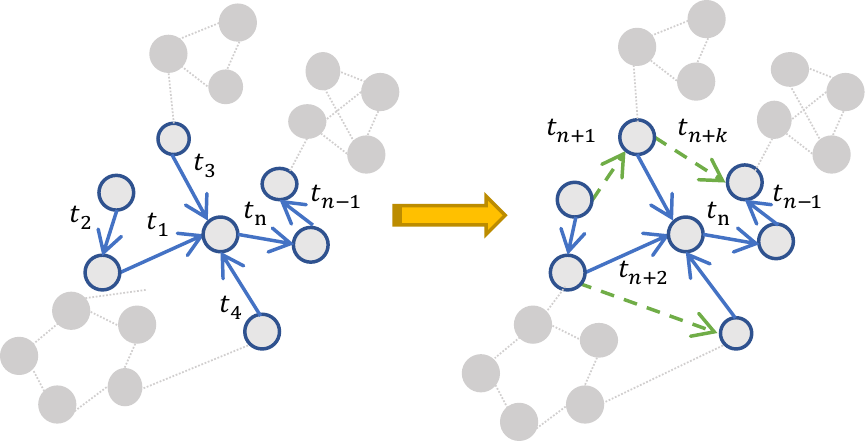}
  \caption{\xh{Community event forecasting on a CTDG: Given a community (nodes and edges with blue strokes) in a historical CTDG, predict where and when the next interaction event (green arrows) will happen. This process can be repeated to forecast to distant future.}
}
  \label{fig:task}
\end{figure}

Continuous Temporal Dynamic Graph (CTDG) is a common representation paradigm for organizing \gq{dynamic} interaction events over time, with edges and nodes denoting the events and the pairwise involved entities, respectively. Each edge also contains information about the timestamp of an event (assuming that the event occurs instantaneously). \xh{For better understanding and forecasting the events in a community, \yh{we propose }the \textbf{community event forecasting} task (Fig.~\ref{fig:task}) in a CTDG 
to predict a series of future events about not only \emph{which} two entities they will involve but also \emph{when} they will occur. \GQd{Use active tense; who proposed this?  We?}\YHd{updated, OK?}}

\yh{\textbf{Problem Formulation. }}Formally, denote a CTDG as $G = (V, E_T)$, where $E_T = \{\varepsilon_i : i = 1, \cdots, T\}$ is the set of edges, $\varepsilon_i = (u_i, v_i, t_i)$ with source node $u_i$, destination node $v_i$, and timestamp $t_i$. The edges are ordered by timestamps, i.e., $t_{i} \leq t_{j}$ given $1 \leq i \leq j \leq T$. We further denote a CTDG within a temporal window as $G_{i:j} = (V, E_{i:j})$ where $E_{i:j} = \{\varepsilon_k : i \le k < j\}$. \xh{Given the queried community (or node candidates that people \yh{are} interested in) $C_q \subset V$,} predicting $K$ future \yh{events within the community }
given $n$ observed events requires to model the following conditional distribution:
\begin{equation}
p(\varepsilon_{n+1}, \cdots, \varepsilon_{n+K} \mid G_{1:n}, C_q)
\end{equation}

where the distribution of each edge $\varepsilon_{n+i}$ is further a triple joint probability distribution of its source and destination nodes as well as its timestamp. 
Compared with traditional time series prediction, event forecasting on a CTDG jointly consider the spatial information characterized by the graph and the temporal signal characterized by the event stream in order to make a more accurate prediction.

There have been several lines of work to approach the problem but all in compromised settings. Graph Neural Networks (GNNs)~\cite{wu2020comprehensive} is an emerging family of deep neural networks for embedding structural information via its message passing mechanism. Temporal Graph Neural Networks (TGNNs) extends GNNs to CTDGs by incorporating temporal signals into the message passing procedure. Among them, Jodie~\cite{Jodie}, TGN~\cite{TGN} and TGAT~\cite{TGAT} are designed for the \emph{temporal link prediction} where the timestamp and one entity of the future event \yh{are} given, i.e., modeling the conditional distribution \xh{$p(v_{n+1} \mid G_{1:n}, u_{n+1}, t_{n+1})$. } Know-Evolve~\cite{Know-Evolve} and DyRep~\cite{DyRep} combine GNNs with methods for multi-variate continuous time series such as Temporal Point Process (TPP) for \emph{event timestamp prediction} which predicts the time of the next event but requires the entities to be known, i.e., modeling the conditional distribution \xh{$p(t_{n+1} \mid G_{1:n}, u_{n+1}, v_{n+1})$}. All these models are not directly applicable to the \xh{community event forecasting problem on CTDGs}.

Another line of work extend TPP methods to incorporate extra signals. Notably, Marked TPP (MTPP) associates each event with a marker and jointly predicts the marker as well as the timestamp of future events. Recurrent Marked Temporal Point Process (RMTPP)~\cite{RMTPP} proposes to use recurrent neural networks to learn the conditional intensity and marker distribution. MTPP and RMTPP are capable of predicting CTDG events by treating the entity pair $(u_i,v_i)$ as the event marker. \xh{However, \yh{these} kinds of MTPP-based methods face three major drawbacks. To begin with, MTPP methods treat edges as individual makers, which are unable to utilize the community and relationship information, resulting in a suboptimal training solution. Besides, individual makers also bring an $O(|V|^2)$ marker distribution space, making model unscalable to giant graphs. The last difficulty is that RMTPP is further constrained by its recurrent structure, which must process each event sequentially for keeping events' contextual correlation. }\YHd{the above two paragraphs seem too detailed and might be better to be placed in the related work section?}

Our contributions in this paper are: 

\textbf{i)} We consider the \xh{community} event forecasting task on a CTDG that jointly predicts the next event's incident nodes and timestamp \xh{within \yh{a certain }community}, which is significantly harder than both temporal link prediction and timestamp prediction. To the best of our knowledge, we are the first who carefully consider and evaluate this task. 

\textbf{ii)} \xh{We handle the \textbf{C}ommunity \textbf{E}vent \textbf{P}redicting task with a graph \textbf{P}oint \textbf{P}rocess model (\textbf{CEP3}), which incorporates both spatial and temporal signals using GNNs and TPPs and can predict event entities and timestamps simultaneously. To scale to large graphs, we factorize the mark distribution of MTPP and reduce the computational complexity from $O(|V|^2)$ of previous TPP attempts~\cite{RMTPP, DyRep} to $O(|V|)$. Moreover, we employ the a time-aware attention model to replace the TPP model's recurrent structure, significantly shortening the sequence length of each training step and enabling training in mini-batches.}


\textbf{iii)} We propose new benchmarks for the \xh{community} event forecasting task on a CTDG. Specifically, we design new evaluation metrics measuring prediction quality of both entities and timestamps. For baselines, we collect and carefully adapt state-of-the-art models from time series prediction, temporal link prediction and timestamp prediction. Our evaluation shows that CEP3 is superior across all four real-world graph datasets.

\section{Related Work}



\subsection{Temporal Graph Learning}
Temporal Graph Learning aims at learning node embeddings using both structural and temporal signals, which gives rise to a number of works. CTDNE~\cite{nguyen2018continuous} and CAWs~\cite{wang2021inductive} incorporate temporal random walks into skip-gram model for capturing temporal motif information in CTDGs. JODIE~\cite{Jodie} and TigeCMN~\cite{DBLP:conf/www/ZhangBEZYL020} adopt \yh{recurrent neural networks (}RNN\yh{s)} and attention-based memory module respectively to update node embedding dynamically. Temporal Graph Neural Networks like TGAT~\cite{TGAT} and TGN~\cite{TGN} enhance the attention-based message passing process from Graph Neural Networks with Fourier time encoding kernel. These attempts focus on the temporal link prediction task. Besides that, other works~\cite{DBLP:journals/tkde/LiZSZLL17,DBLP:conf/kdd/WuGGWC19} focus on information diffusion task which aims at predicting whether a user will perform an action at time $t$. None of them is designed for event forecasting.
RE-Net~\cite{RE-Net} and CoNN~\cite{CoNN} study a similar event forecasting setting on Discrete Temporal Dynamic Graphs (DTDGs). However, continuous time prediction is much harder and their methods cannot be directly applied.

\subsection{Temporal Point Process}
\label{sec:tpp}

A temporal point process (TPP) \cite{TPP} is a stochastic process modeling the distribution of a sequence of events associated with continuous timestamp\yh{s} $t_1, \cdots, t_n$.  A TPP is mostly characterized by a conditional intensity function $\lambda(t)$, from which it computes the conditional probability of an event occurring between $t$ and $t+dt$ given the history $\{t_i : t_i < t\}$ as $\lambda(t)dt$.  According to \cite{aalen2008survival}, the log conditional probability density of an event occurring at time $t$ can be formulated as
\begin{equation}
\label{eqn:logprob-time}
f(t) = \log \lambda(t) - \int_{t_n}^t \lambda(\tau) d\tau
\end{equation}

Additionally, a marked temporal point process (MTPP) associates each event with a \emph{marker} $y_i$ which is often regarded as the \emph{type} of the event.  MTPP thus not only models when an event would occur, but also models what type of event it is.  Event forecasting over CTDGs can also be modeled as a MTPP if treating the event's incident node pair as its marker.  The conditional intensity of the entire MTPP can be modeled as a sum of conditional intensities of each individual marker: $\lambda = \sum_{m} \lambda_m$.
This allows it to first make the prediction of event time, and then predict the marker conditioned on time via sampling from a categorical distribution: $m \sim \text{Categorical}(\lambda_m)$. This avoids modeling time and marker jointly. Such idea has been widely used in follow-up works such as Recurrent Marked Temporal Point Process (RMTPP)~\cite{RMTPP}. 
RMTPP parametrizes the conditional intensity and the marker distribution with a recurrent neural network (RNN). RMTPP's variants include CyanRNN~\cite{CyanRNN} and ARTPP~\cite{ARTPP}.


We also demonstrate several other relevant works and applications related to TPPs~\cite{DBLP:conf/ijcai/ShchurTJG21}.
CoEvolving~\cite{Coevolving}, a variant model of MTPP, uses Hawkes processes to model the user-item interaction, respectively. NeuralHakwes ~\cite{NeuralHakwes} relaxes the positive influence assumption of the Hawkes process by introducing a self-modulating model.
DeepTPP~\cite{DeepTPP} models the event generation problem as a stochastic policy and applied inverse reinforcement learning to efficiently learn the TPP. 
However, these methods cannot be directly applied to event forecasting on CTDGs due to the following reasons.
First, a CTDG is essentially a \emph{single} event sequence, whose length ranges from tens of thousands to hundreds of millions. RNNs are known to have trouble dealing with very long sequences.  One may consider dividing the sequence into multiple shorter windows, which will make the events disconnected within the given window and discard all the data before it.  This fails to explore the dependencies between events that are distant over time but topologically connected (i.e. sharing either of the incident nodes).  One may also consider training an RMTPP with Truncated BPTT \cite{jaeger2002tutorial} on the long sequence as a whole. \yh{However,} this is inefficient because parallel training is impossible due to its recurrent nature, \yh{which means that} one has to unroll the sequence one event at a time.  Second, directly modeling the marker generation distribution with a\yh{ unit} softmax\yh{ function} will produce \yh{a vector with space complexity of} $O(|V|^2)$, as the event markers will be essentially the events' incident node pairs. This is undesirable for large graphs.

\cite{farajtabar2017coevolve} models link and retweet generation on a social network with a TPP, and also provide\yh{s} a simulation algorithm that generates from\YHd{remove ``from"? ``a simulation algorithm that generates the TPP model" or ``a simulation algorithm generated from the TPP model"?}\yh{ the} TPP model. It is very similar to our task except that it is focused on a specific social network setting, and \yh{the authors} did not quantitatively evaluate the quality of \yh{the} simulation model.

\subsection{Temporal Point Process on Dynamic Graph}
\label{sec:TPPonG}
Previous works, such as \cite{DyRep, 9138462, DBLP:journals/tit/HallW16}, use kinds of recurrent architecture to approximate temporal point process over graphs. However, recurrent architecture prevents the model from parallelized minibatch training, which is undesirable especially on large-scale graphs. \yh{This is because} learning long-term dependencies using recurrent architecture requires the model to traverse the event sequence one by one instead of randomly mini-batch selection. 

MMDNE~\cite{DBLP:conf/cikm/LuWSYY19}, HTNE~\cite{DBLP:conf/kdd/ZuoLLGHW18} and DSPP~\cite{DSPP} \yh{employ the} attention mechanism to 
avoid the inefficiency of the recurrent structure in training with large CTDG\yh{s}. However, these works are\yh{ restrictively} dedicated to link prediction or timestamp prediction task. Adapting the two models to event forecasting requires non-trivial changes since neither of them handles efficient joint forecasting of the event's incident nodes.

\section{Model}
\label{sec:Method}
\begin{figure}[tb!]
\centering
  \includegraphics[width=\linewidth]{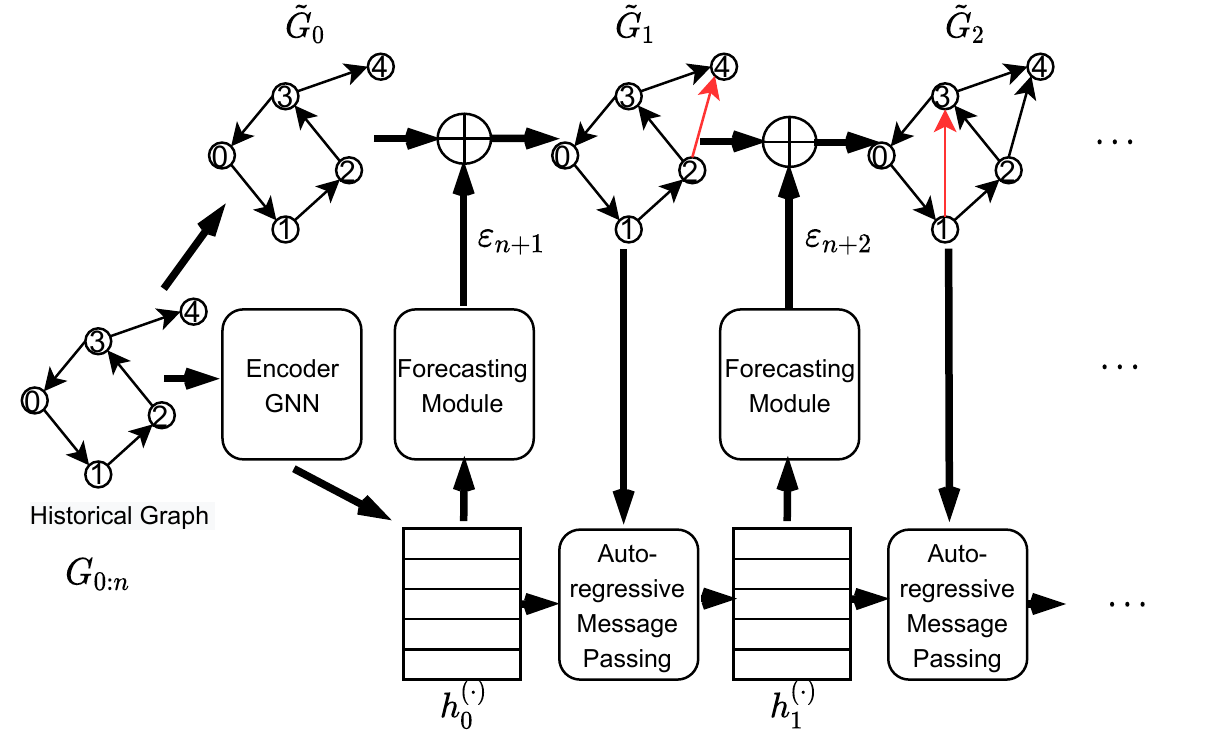}
  \caption{The overall architecture of proposed CEP3 model.  Red arrows represent the predicted events $\varepsilon_{n+i}$.  
  \YHd{Should this figure be removed as we have a new diagram?}}
  \label{fig:model}
\end{figure}

Our model is depicted in Fig.~\ref{fig:model}.  To predict the next $K$ events $\varepsilon_{n+1}, \cdots, \varepsilon_{n+K}$ given the history graph $G_{1:n}$, we first obtain an initial representation $h_0^{(\cdot)}$\YHd{do we need a bracket for the cdot?} for every node using an GNN Encoder, as well as an initial graph $\tilde{G}_0$.  Then for the $i$-th step, we predict $\varepsilon_{n+i}=(u_{n+i}, v_{n+i}, t_{n+i})$, i.e. the source node, destination node, and timestamp for the next $i$-th event.  The predicted event is then added into $\tilde{G}_{i-1}$ to form $\tilde{G}_i$, to keep track of what we\yh{ have} predicted so far.  The hidden states $h_i^{(\cdot)}$ are then updated from the new graph $\tilde{G}_i$ and $h_{i-1}^{(\cdot)}$\YHd{again, do we need a bracket for the cdot?}.  This generally follows the framework of RMTPP \cite{RMTPP}, except that \textbf{i)} we initialize the recurrent network states with a time-aware GNN, which allows our recurrent module to traverse over a much shorter sequence without losing historical information, \textbf{ii)} we update the recurrent network states with a GNN to model the topological dependencies between entities caused by new events, and \textbf{iii)} we forecast the nodes and \yh{the }timestamp\yh{ for an event} by decomposing the joint probability distribution.  We give specific details of each component as follows.


\subsection{Structural and Temporal GNN Encoder}
\label{sec:Encoder}
The Encoder GNN should be capable of encoding relational dependencies, timestamps, and optionally edge features at the same time.  Therefore, it has the following form:
\begin{multline}
\label{eqn:encoder}
z_l^{(v)} = f_\text{agg}^\text{init}(\{g_\text{msg}^\text{init}(z_{l-1}^{(u)}, z_{l-1}^{(v)}, e, \phi(t_n - t)) \\ 
\qquad : (u,v,t,e) \in \mathcal{N}^v[0:n]\})
\end{multline}
where $\mathcal{N}^v[0, n]$ represents the neighborhood of $v$ on the graph $G_{1:n}$ and $\phi(t)$ are learnable time encodings used in \cite{TGAT,TGN,Time2Vec}. $f_\text{agg}^\text{init}$ and $g_\text{msg}^\text{init}$ can be any aggregation and message functions of a GNN.  $z_0^{(v)}$ could be either node $v$'s feature vector if available or a zero vector otherwise.  We then initialize $h_0^{(v)} = z_L^{(v)}$.

We use a GNN to initialize the recurrent network's states because it takes all the historical events within a topological local neighborhood, including the incident nodes, the timestamps, and the feature of events together\yh{,} as input, while enabling us to train on multiple history graphs in parallel.  In particular, we use neighborhood graph temporal attention based method for encoding, whose detailed formulation is as follows.
Temporal Graph Attention Module is a self attention based node embedding method inspired by \cite{TGAT}, the detailed formulation is as below:
\begin{equation}
\begin{aligned}
    \textbf{z}^{(l),t}_i =& \texttt{MLP}(\textbf{z}^{(l-1),t}_i||\tilde{\textbf{z}}_i^{(l),t})\\
    \tilde{\textbf{h}}^{(l),t}_i =& \texttt{MultiHeadAttn}^{(l)}(q^{(l),t}_i,K^{(l),t}_i, V^{(l),t}_i)\\
    \textbf{q}^{(l),t}_i =& [z^{(l-1),t}_i||\phi(0)]\\
    \textbf{K}^{(l),t}_i =& \textbf{V}^{(l),t}_i = \textbf{C}^{(l),t}_i\\
    \textbf{C}^{(l),t}_i =& [h_j^{(l-1),t}||e_{ij}^{t_j}||\phi(t_\text{src}-t_j),j\in \mathcal{N}]
\end{aligned}
\end{equation}
For each source node at time $t_\text{src}$, we sample its two hop neighbors whose timestamp is prior to $t_\text{src}$. The multi-head attention is computed as:
\begin{equation}
    \begin{aligned}
        \tilde{\textbf{z}}_{(l)}^t = \sum_{a=0}^n\text{SoftMax}\left(\frac{(W_{Q,a}^{(l)}\textbf{q}_i^{(l),t})(W_{K,a}^{(l)}\textbf{K}_i^{(l),t})}{\sqrt{d_k}}\right)\left(W_{V,a}^{(l)}\textbf{V}_i^{(l),t}\right)
    \end{aligned}
\end{equation}
The temporal encoding module is the same as in \cite{TGAT, TGN} original paper:
\begin{equation}
\begin{gathered}
    \phi(\Delta{t}) = \cos(\vec\omega{t}+\vec b)
\end{gathered}
\end{equation}
Where $\omega$ and b are learnable parameters. 
\XHd{In TGAT~\cite{TGAT} K-hop neighbor nodes are sampled recursively from the original source nodes. For example, k hop's neighbors with respect to k-1 hop's source nodes are required to have timestamp smaller than those source nodes in k-1 hop. Such sampling scheme is inefficient and lack of ability to consider all history event happened before the original node's query timestamp $t_\text{src}$. We use a simpler scheme by directly sample k-hop neighbors whose timestamps are prior to original source nodes.}

Although a GNN cannot consider events outside the neighborhood, we argue that such\yh{ an} impact is minimal. \xh{We empirically verify our belief by comparing against the variant that uses both attention and RNN based memory module in the training phase (named \textbf{CEP3 w RNN}), which can incorporate historical events and time-aware information by the view of topological locality and recursive impact, respectively. However, RNN takes drastically more memory and time in training because of the same reason in Section~\ref{sec:TPPonG}.}

\subsection{Hierarchical Probability-Chain Forecaster}

\begin{figure*}[tb!]
\centering
 \includegraphics[width=0.8\textwidth]{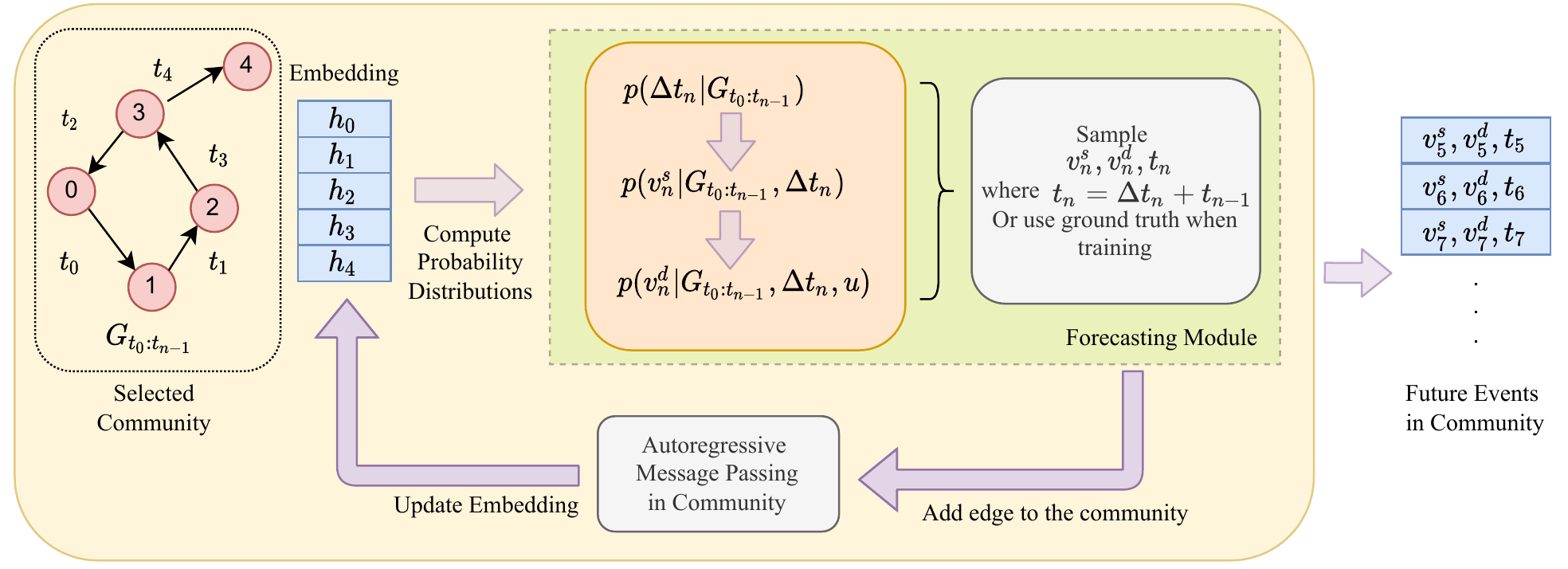}
 \caption{\xh{The hierarchical probability-chain forecaster and its workflow relationship with the auto-regressive message passing module. The node embeddings are learned from the GNN Encoder described in Section \ref{sec:Encoder}.  Note that the `selected community' refers to an application-dependent collection of candidate nodes for query.}\YHd{perhaps make the figure larger if we have extra space?}}
 \label{fig:Forecaster}
\end{figure*}

\label{sec:forecasting}
\xh{Fig.~\ref{fig:Forecaster} demonstrate the detail\yh{s} of our event forecaster. We can see that the forecaster predicts future events only according to node embeddings and historical connections in the selected candidates (or communities). It means that we do not have to be concerned about a large number of communities \yh{which are likely to slow }
down the process, because CEP3 model can handle numerous communities simultaneously during training and inference. }

From the superposition property\cite{superposition} of an MTPP described in Section~\ref{sec:tpp}, we forecast the next event $\varepsilon_{n+i}=(u_{n+i}, v_{n+i}, t_{n+i})$ by a triple probability-chain:
\begin{multline}
p(u_{n+i}, v_{n+i}, t_{n+i}) = p(t_{n+i}) \\
p(u_{n+i} \mid t_{n+i}) p(v_{n+i} \mid t_{n+i}, u_{n+i})
\end{multline}
It means that we can predict first the timestamp, then the source node, and finally the destination node.  We predict $t_{n+i}$ by modeling the distribution of the time difference $\Delta t_{n+i} = t_{n+i} - t_{n+i-1}$ as follows:
\begin{equation}
\begin{gathered}
\lambda^{(v)}_i = \mathrm{Softplus}(\mathrm{MLP}_\text{t}( h_{i-1}^{(v)})) \\
\lambda_i = \sum_{v} \lambda^{(v)}_i \\
\Delta t_{n+i} \sim  Exponential(\lambda_i) 
\end{gathered}
\end{equation}
\xh{where $\mathrm{Softplus}$ ensures that $\lambda_i^{(v)}$ is above zero and the gradient still exists for negative $\lambda_i^{(v)}$\yh{ values}. $\mathrm{MLP}_t$ represents a multilayer perceptron to generate \yh{the }intensity value of time. Since $\Delta t_{n+i}$ obeys exponential distribution, we simply sample the mean value of time intensity distribution, $\frac{1}{\lambda_i}$, as the final $\Delta t_{n+i}$ output \yh{during }
training.} 
The conditional intensity of the next event occurring on \emph{one} of the nodes in $V$ is thus the sum of all $\lambda_i^{(v)}$ \cite{superposition}.  Other conditional intensity choices are also possible.

We then generate the source node $u_{n+i}$ of event $\varepsilon_{n+i}$ from a categorical distribution conditioned on $t_{n+i}$, parametrized by another MLP:
\begin{equation}
\label{eqn:u}
p(u_{n+i}) = \mathrm{Softmax}(\mathrm{MLP}_\text{src}(h_{i-1}^{(u)} \Vert \phi(\Delta t_{n+i})))
\end{equation}
where $\Vert$ means concatenation and $\phi$ has the same form as in Eq.~\ref{eqn:encoder}.
We then generate the destination node $v_{n+i}$ similarly, conditioned on $t_{n+i}$ and $u_{n+i}$:
\begin{equation}
\label{eqn:v}
p(v_{n+i}) = \mathrm{Softmax}(\mathrm{MLP}_\text{dst}(h_{i-1}^{(v)} \Vert h_{i-1}^{(u_{n+i})} \Vert \phi(\Delta t_{n+i})))
\end{equation}

Note that the formulation above will only generate two distributions that have $|V|$ elements, instead of $|V|^2$ as in RMTPP\yh{ \cite{RMTPP}}.  The implication is that during inference the strategy will be \emph{greedy}: we first pick whatever source node that has the largest probability, then we pick the destination node conditioned on the picked source node.  To verify the impact of this design choice, we also explored a variant of our method where we generate a joint distribution of the pair $(u_{n+i}, v_{n+i})$ with $O(|V|^2)$ elements, which we name \textbf{CEP3 w/o HRCHY} (short for Hierarchy).\YHd{Would it be more realistic to pick the pair once instead of source and target one after the other? What prevent us from doing so? Complexity?}

If a dataset has a lot of nodes, evaluating $p(u_{n+i})$ and $p(v_{n+i})$ will incur a linear projection with $O(|V|)$ complexity.  This is another obstacle of scaling up to larger datasets. 
 

\subsection{Auto-Regressive Message Passing}
As shown in Fig.~\ref{fig:Forecaster}, we assume that an event's occurrence will directly influence the hidden states of its incident nodes.  Moreover, the influence will propagate to other nodes along the links created by historical interactions. Therefore, after generating the new event $\varepsilon_{n+i}$, we would like to update the nodes' hidden states by message passing on the graph with the new events.  We achieve that by maintaining another graph $\tilde{G}_i$ that keeps track of the graph with the historical interactions $G_{1:n}$ and the newly predicted events up to $\varepsilon_{n+i}$.

Specifically, we initialize $\tilde{G}_0$ with the candidate node set $C$ as its nodes.  Two candidate nodes are connected in $\tilde{G}_0$ if their distance is within $L$ hops.  The resulting graph encompasses the dependency between candidate nodes during the encoding stage.  Every time a new event $\varepsilon_{n+i}$ is predicted, we add the event back in: $\tilde{G}_i = \tilde{G}_{i-1} \cup \varepsilon_{n+i}$.

Afterwards, we update the nodes' hidden states using a message passing network such as GCN~\cite{GCN} for spatial propagation and a GRU~\cite{cho-etal-2014-learning} for temporal propagation:
\begin{equation}
    \begin{aligned}
    w_{i,0}^{(v)} &= h_{i-1}^{(v)} \\
    w_{i,l}^{(v)} &= f_{\text{agg}}^{\text{upd}}(\{g_\text{msg}^{\text{upd}}(w_{i,l-1}^{(u)} , w_{i,l-1}^{(v)}) : u\in \mathcal{N}_{\tilde{G}_{i}}^v \}) \\
    h_i^{(v)} &= \mathrm{GRU}(\left[ w_{i,P}^{(v)} \Vert \phi(\Delta t_{n+1})\right], h_{i-1}^{(v)})
    \end{aligned}
\end{equation}
where $\mathcal{N}_{\tilde{G}_i}^v$ is the neighboring events of node $v$ in $\tilde{G}_i$, $f_{\text{agg}}^{\text{upd}}$ can be any message aggregation function and $g_\text{msg}^{\text{upd}}$ can be any message function.

To verify the necessity of updating the community using message passing after a event, we also explore a variant where we do not update all the node's hidden states in the community, but only the incident nodes $u_{n+i}$ and $v_{n+i}$. We name this variant \textbf{CEP3 w/o AR}.

\subsection{Loss Function and Prediction}
The forecasting module outputs the next event's timestamp $t_{n+i}$ and indicent nodes $u_{n+i}$ and $v_{n+i}$ for all events $\varepsilon_{n+i}$, which\YHd{what does ``which" refer to?} are minimized via negative log likelihood.  Specifically, the loss function goes as follows:
\begin{multline}
    \mathcal{L}_\text{time} = \sum_{i=1}^K [ \underbrace{-\log(\lambda_i) + \Delta t_{n+i} \lambda_i}_{\text{time loss}}  \\
    \underbrace{-\log p(u_{n+i}) -\log p(v_{n+i})}_{\text{entity loss}} ]
\end{multline}
where the first two terms within summation are log survival probability from Eq. \ref{eqn:logprob-time} and the last two terms are log probabilities for source and destination node prediction. The time integration term $\int_{t_n}^t \lambda(\tau) d\tau$ in Eq. \ref{eqn:logprob-time} is approximated using first order integration method by $\lambda(t)\Delta t$ for the ease of computation. \XHd{TBD: Add more description about loss function}

\section{Experiments}
\subsection{Datasets}
\label{apd:datasets}
In this section, we test the performance and efficiency of the proposed method against multiple baselines on four public real-world temporal graph datasets: Wikipedia, MOOC \cite{Jodie}, GitHub \cite{DyRep}, and SocialEvo \cite{madan2011sensing}.  Table~\ref{tab:datasets} shows \yh{summary} statistics of the datasets used in our experiments. \yh{A} detailed description is put in the below.    

\begin{table}[t!]
\caption{Statistics of the datasets used in our experiments.}
\centering
\small
\resizebox{\linewidth}{!}{ 
\begin{tabular}{@{}clrrrr@{}}
\toprule
Level & Statistics                   & Wikipedia      & MOOC    & Github      & SocialEvo              \\ \midrule
\multirow{9}{*}{\makecell[c]{Graph\\level}} & Edges                        & 157,474         & 411,749 & 20,726       & 62,009                \\
&Nodes                        & 9,227           & 7,145    & 282         & 83                     \\
&Max Degree          & 1,937           & 19,474    & 4,790         & 15,356 \\
&Aver. Degree          & 34           & 115    & 147         & 1,310 \\
&Edge Feat. Dim.             & 172            & 4       & 10          & 10             \\
&Is Bipartite                 & True           & True   & False       & False                    \\
&Timespan                     & 31days         & 30days        & 1years      & 74days   \\
&Edges/hour & 211.66         & 576.30   & 2.36       & 7.79             \\
&Data Spilt                   & \multicolumn{4}{c}{70\%-15\%-15\% by timestamp order}       \\ \midrule
\multirow{9}{*}{ \makecell[c]{Community\\level}} & Communities       & 142       &25   & 17        & 10             \\
&Max Nodes  & 396          & 990   & 46       & 18             \\
&Aver. Nodes  & 50.27          & 264.96   & 15.94       & 7.7             \\
&Max Edges  & 4799         & 11686   & 3221       & 12199             \\
&Aver. Edges & 778.28          & 2560.00   & 534.71       & 3420.90             \\
&Min Edges & 77          & 16   & 34       & 863             \\
&Max Edges/hour & 6.47          & 33.33   & 0.36       & 1.99             \\
&Aver. Edges/hour & 1.11         & 5.36   & 0.06       & 0.56             \\
&Min Edges/hour & 0.14          & 0.34   & 0.01       & 0.15             \\
 \bottomrule
\end{tabular}
}
\label{tab:datasets}
\end{table}
\normalsize

\textbf{Wikipedia}
~\cite{Jodie} dataset is widely used in temporal-graph-based recommendation systems.  It is a bipartite graph consisting of user nodes, page nodes and edit events as interactions.  
We convert the text of each editing into a edge feature vector representing their LIWC categories~\cite{LIWC}.


\textbf{MOOC}
~\cite{Jodie} dataset, collected from a Chinese MOOC learning platform XuetangX, consists of students' actions on MOOC courses, e.g., viewing a video, submitting an answer, etc. 

\textbf{Github}
~\cite{DyRep} dataset is a social network built from GitHub user activities, where all nodes are real GitHub users and interactions represent user actions to the other's repository such as Watch, Fork, etc. 
Note that we do not use the interaction types as we follow the same setting as \cite{DyRep}.

\textbf{SocialEvo}
~\cite{DyRep, madan2011sensing} dataset is a small social network collected by MIT Human Dynamics Lab.

Since the public dataset SocialEvo and Github have no edge feature, we generate a 10-dimensional edge feature using following attributes, including the current degrees of the two incident nodes of an edge, and the time differences between current timestamp and the last updated timestamps of the two incident nodes. Note that the time differences are described in the numbers of days, hours, minutes and seconds, respectively.


\subsection{Baselines}
\label{apd:baselines}
\begin{table*}[t!]
\caption{Comparison of model capabilities. Note that the usage of RNN prevents \yh{a }model from parallel training as is mentioned in Section~\ref{sec:TPPonG}.  
$^*$Requires non-trivial adaptation.\YHd{we do not have appendix now, please update the caption!}}
\centering
\small
\begin{tabular}{@{}lcccccccc@{}}
\toprule
Taxnomy               & \multicolumn{2}{c}{GNN+TPP}  & {RNN+TPP}    & GNN    &   \multicolumn{2}{c}{TPP} \\ \midrule
Methods                                & CEP3    & DyRep    &  RMTPP      & TGAT       & Poisson & Hawkes  \\ \midrule
Predicts Link (u,v)             &  $\surd$   & $\surd$       &   $\surd$    & $\surd$ & $\surd$ & $\surd$ \\
Predicts Continuous Time t      &   $\surd$ & $\surd$       &   $\surd$  &          &         $\surd$ & $\surd$          \\
Jointly Predicts Event (u,v,t)  &   $\surd$ & $\surd^*$     &   $\surd$  &                   & $\surd$ & $\surd$         \\ 
Explicitly Models Topological Dependency      & $\surd$   &  $\surd$ &            &  $\surd$   &         &                           \\
Complexity of Node Prediction   & $O(|V|)$   & $O(|V|^2)$    &      $O(|V|^2)$        &   $O(|V|^2)$   &    $O(|V|^2)$ & $O(|V|^2)$ \\
Parallel Training               & $\surd$   &                        &            &$\surd$   &          $\surd$ &   \\
Captures Sequential Info with       & Attention & RNN+Attention    &    RNN     &Attention     & Poisson Process      & Hawkes Process  \\ 
\bottomrule
\end{tabular}
\label{tab:baselines}
\end{table*}

In addition to \textbf{CEP3}, \textbf{CEP3 w RNN}, \textbf{} and \textbf{CEP3 w/o AR} mentioned in Section~\ref{sec:Method},
we compare against the following baselines: a Seq2seq model with a \textbf{GRU} \cite{cho-etal-2014-learning}, a Poisson Process (\textbf{TPP-Poisson}), a Hawkes Process (\textbf{TPP-Hawkes})~\cite{hawkes1971spectra}, \textbf{RMTPP}~\cite{RMTPP} and its variant with the same two-level hierarchical factorization as in Eq. \ref{eqn:u} and \ref{eqn:v} (\textbf{RMTPP w HRCHY}), an adaptation of DyRep \cite{DyRep} and an auto-regressive variant (named \textbf{DyRep} and \textbf{DyRep w AR}). 
Notably, RMTPP and its variants are SOTA models for MTPPs in general, and DyRep is SOTA in temporal link prediction and time prediction on CTDG.  Since our task is new, we made adaptations to the baselines above, with details of each baseline is as follows:

\textbf{Time series Methods}:
For baseline model of sequential prediction, we build an RNN model, Gated Recurrent Unit (GRU). Each source and destination cell has a hidden state, the output of the model will be predicted time mean and variance as well as probability for each class to interact, the time will be formulated as Gaussian distribution and source and destination node will be formulated as categorical distribution. This formulation forces GRU predicting timestamp of upcoming events only depending on the hidden state in RNN, whereas other baselines adapt TPP function as a stochastic probability process, obtaining a better modeling capability.
We use a \textbf{GRU} \cite{cho-etal-2014-learning} as a simple baseline without using TPP to model time distribution, treating event forecasting on CTDG as a sequential modeling task.  It takes in the event sequence and outputs the next $K$ event in a Seq2seq fashion \cite{sutskever2014seq2seq}.  The loss term for time prediction is mean squared error and the loss term for source and destination prediction is the negative log likelihood.  This formulation forces GRU to predict the timestamp of upcoming events only depending on the hidden state in RNN, whereas other baselines adapt TPP for better modeling capabilities.

\textbf{Temporal Point Process Methods}:
Following the benchmark setting of \cite{RMTPP}, we compared our model against other traditional TPP models and deep TPP models.
\begin{itemize}
\item \textbf{TPP-Poisson}: 
We assume that the events occurring at each node pair $(u, v)$ follows a Poisson Process with a constant intensity value $\lambda_{u,v}$, which are learnt from data via Maximum Likelihood Estimation (MLE).
\item \textbf{TPP-Hawkes}~\cite{hawkes1971spectra}: 
We assume that the events occurring at each node pair $(u, v)$ follows a Hawkes Process with a base intensity value $\mu_{u,v}$ and an excite parameter $\alpha_{u,v}$, which are learnt from data via MLE.
\item \textbf{RMTPP}~\cite{RMTPP}: We directly consider each source and destination node pair as a unique marker.  We note that this formulation will exhaust memory and time on graphs with more than a few thousand nodes, since RMTPP will assign a learnable embedding for each node pair, resulting in $O(|V|^2)$ (Here $V$ is total number of nodes in the entire CTDG) parameters which is too expensive to update.
\item \textbf{RMTPP w HRCHY}: We consider a variant of RMTPP where we replace the source-destination node prediction with our hierarchical formulation: we first select the source node, then condition on the source node we select the destination node. The latter formulation can also serves as an ablation study to demonstrate the usage of considering the graph structure. 
\item \textbf{DyRep}~\cite{DyRep}: To the best of our knowledge, DyRep is the most popular work that combine the temporal point process with graph learning techniques to model both temporal and spatial dependencies.  
Since the original DyRep formulation only handled temporal link prediction and time prediction, but not autoregressive forecasting, we compute an intensity value $\lambda_{u,v}$ with DyRep for each node pair and assumed a Poisson Process afterwards.
\item \textbf{DyRep w AR}: We made a trivial adaptation to the original formulation of DyRep by updating the source and destination node involved once a new edge is added to the graph.  The update function is identical to DyRep updating function during embedding.  
This benchmark is designed to demonstrate that the propagation from newly forecast event to local neighborhood is necessary in getting better performance.
\end{itemize}

We also crave our proposed model for having the ability to implement minibatch training. As described in the beginning of Section~\ref{sec:Method}, our CEP3 achieves large-scale and parallelized training by utilizing Hierarchical TPP and GNN based updating module, respectively. \yh{A} summary of\yh{ the} mentioned baselines is shown in Table~\ref{tab:baselines}, the proposed model CEP3 satisfies all the desirable properties.

\subsection{Configuration}
\label{apd:config}
We provide the details of our network architecture, the hyper-parameters and the selected community detection method for better reproducibility.
Table~\ref{table:config} summarized other key parameters in our model. From hardware perspective, for all the experiments we train our model and benchmark models on Intel(R) Xeon(R) Platinum 8375C CPU @ 2.90GHz. 
\yh{Our code is based on PyTorch and Deep Graph Library~\cite{wang2019dgl}, and will be published after acceptation.}
\begin{table}[t]
\caption{Configurations for our CEP3 and all baselines.}
\centering

 \begin{tabular}{c c} 
 \toprule
 Name & Value \\ [0.5ex] 
 \midrule
 Hidden Dim in Encoder & 100 \\ 
 
 Hidden Dim in Forecaster & 50 \\
 
 Community Dim & 10 \\
 
 Layers in MLPs & 2 \\
 
 K-hops & 2 \\
 
 Sampled neighbors/Hop & 15 \\
 
 Learning rate & 0.0001\\
 
 Optimizer & Adam\\
 
 \# of attn head & 4\\
 
 Recurrent Module & GRU \\
 
 Epochs & 100 \\
 
 Forecasting Window & 200 Steps \\
 
 Community Detection Method & Louvain\\
 \bottomrule
\end{tabular}
\label{table:config}
\end{table}

\subsection{Evaluation Metrics}
\xh{One of the most important contributions of our study is the concept of community event prediction, which is a novel task in the field of temporal graph\yh{s}. To the best of our knowledge, we are the first 
\yh{to }consider forecasting events 
\yh{sampled }from a temporal point process model, given a historical graph and a candidate node set in which \yh{the }user is interested. For 
\yh{evaluation on }a specific dataset, we utilize the communities segmented by the conventional community detection algorithm Louvain~\cite{Louvain} as the candidate node set, and we report the average result of all communities.   }

For each community $C_q$, we measure the perplexity ($PP_{C_q}$) of the ground truth source and destination node sequence for evaluating the node predicting performance, and evaluate the mean absolute error ($MAE_{C_q}$) of the predicted timestamps. Using our MAE to evaluate the quality of auto-regressive forecasted sequence over multiple timesteps can also be seen in traffic flow prediction \cite{dcrnn}. 



\xh{Perplexity (PP) \cite{Perplexity} is a concept in information theory that assesses how closely a probability model's projected outcome matches the real sample distribution. The less perplexity the situation, the higher the model's prediction confidence. In the field of natural language processing, \yh{p}erplexity is also commonly used to evaluate a language model's quality, i.e., to evaluate how closely the sentences generated by the language model match real human language samples. \yh{A language }
model predict\yh{s the} next word from the word dictionary, whereas our event predicting model select\yh{s} nodes from the node candidates (community). Therefore, it 
\yh{is reasonable to employ }perplexity as a metric in our task.
}


Specifically, suppose we have the communities' ground truth event sequence $(u_i, v_i, t_i)$ and the prediction sequence $(\hat{u}_i, \hat{v}_i, \hat{t}_i)$ where $i=1, \cdots, K$.  For we compute per step PP as
\begin{equation}
\small
PP = \exp {\left(-\frac{1}{K}\sum_{i=1}^K \left[ \log p(u_{i}) + \log p(v_{i} \mid u_{i})\right]\right)}
\end{equation}
In order to measure the distance between two sequence with difference length, we compute MAE following \cite{xiao2017wasserstein} as:
\begin{equation}
MAE = \frac{1}{K(t_K - t_0)} \sum_{i=1}^K \left[ \left| t_{i} - \min(t_{K}, \hat{t}_{i}) \right| \right]
\end{equation}
\xh{To keep it comparable in diverse datasets, the MAE is divided by the max time \yh{span} $t_K - t_0$ and the sequence length $K$. We report the average $PP_{C_q}$ of all communities as the $PP$ of a certain dataset, and $MAE$ is calculated in the same way. Smaller values of both metrics indicate the better model performance.}



\subsection{Result Analysis}

From Table~\ref{tab:eventforecastresult} we can see the obvious advantage of CEP3 compared to other baselines in different datasets under both MAE and perplexity. The MAE difference between GRU and RMTPP show the effectiveness of temporal point process in predicting timestamps. Comparing CEP3 with sequence based TPP models RMTPP, we can see that using GNN to capture historical interaction information can improve the forecasting performance. Further more, when comparing \textbf{DyRep w AR} versus \textbf{DyRep} and \textbf{CEP3 w/o AR} versus pure \textbf{CEP3}, we can conclude that auto-regressive updates can better capture the impact of newly predicted events.

Our model performs better than DyRep w AR because in our CEP3, during the auto-regressive update, the newly predicted event not only influences the node involved in the event but also propagates to other nodes via message passing. 

\subsection{Forecasting Visualization}
\begin{figure*}[!tbh]
	\centering
	\subfigure[Ground Truth.]{
		\begin{minipage}[b]{0.25\linewidth}
			\includegraphics[width=\linewidth]{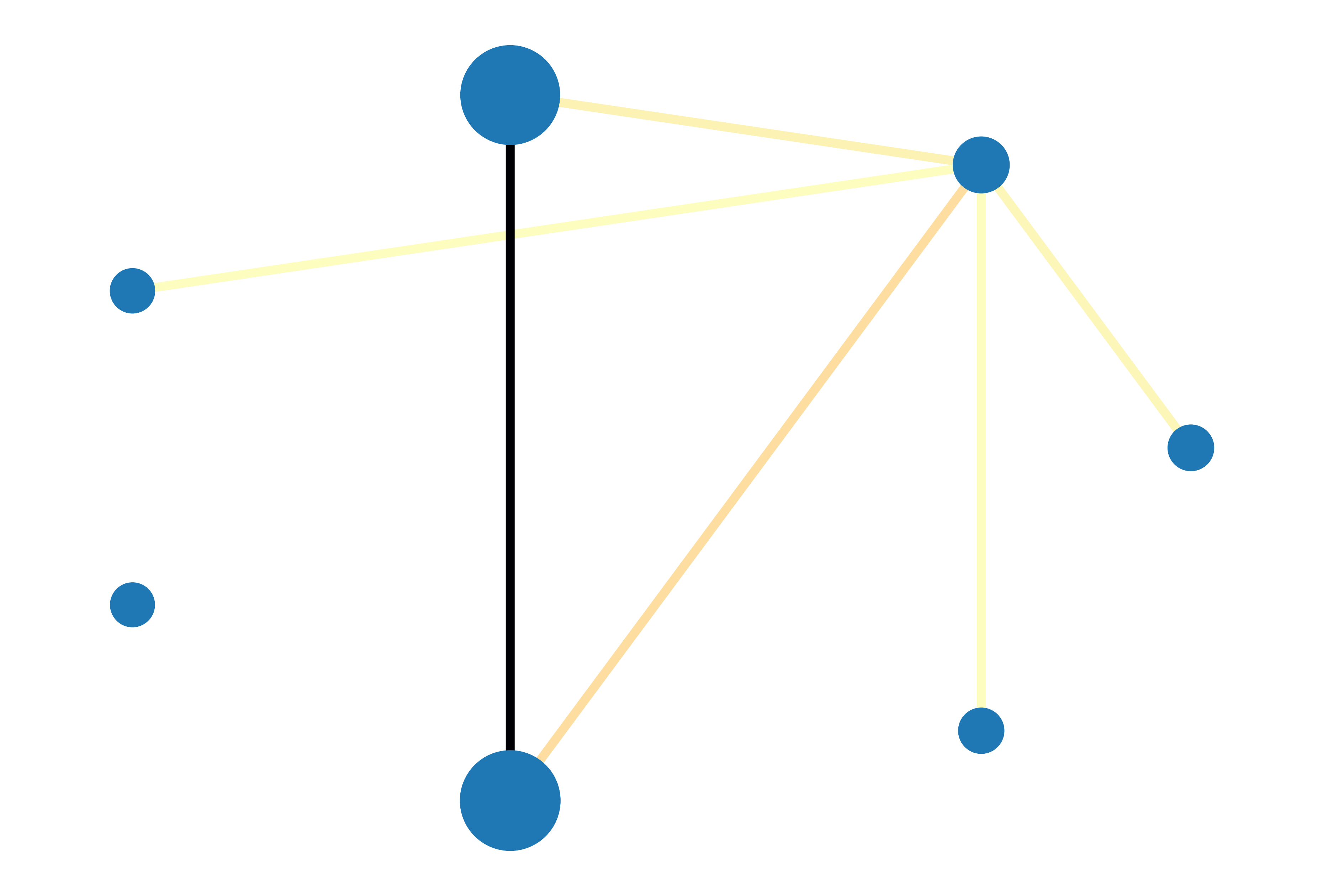}
			
			\includegraphics[width=\linewidth]{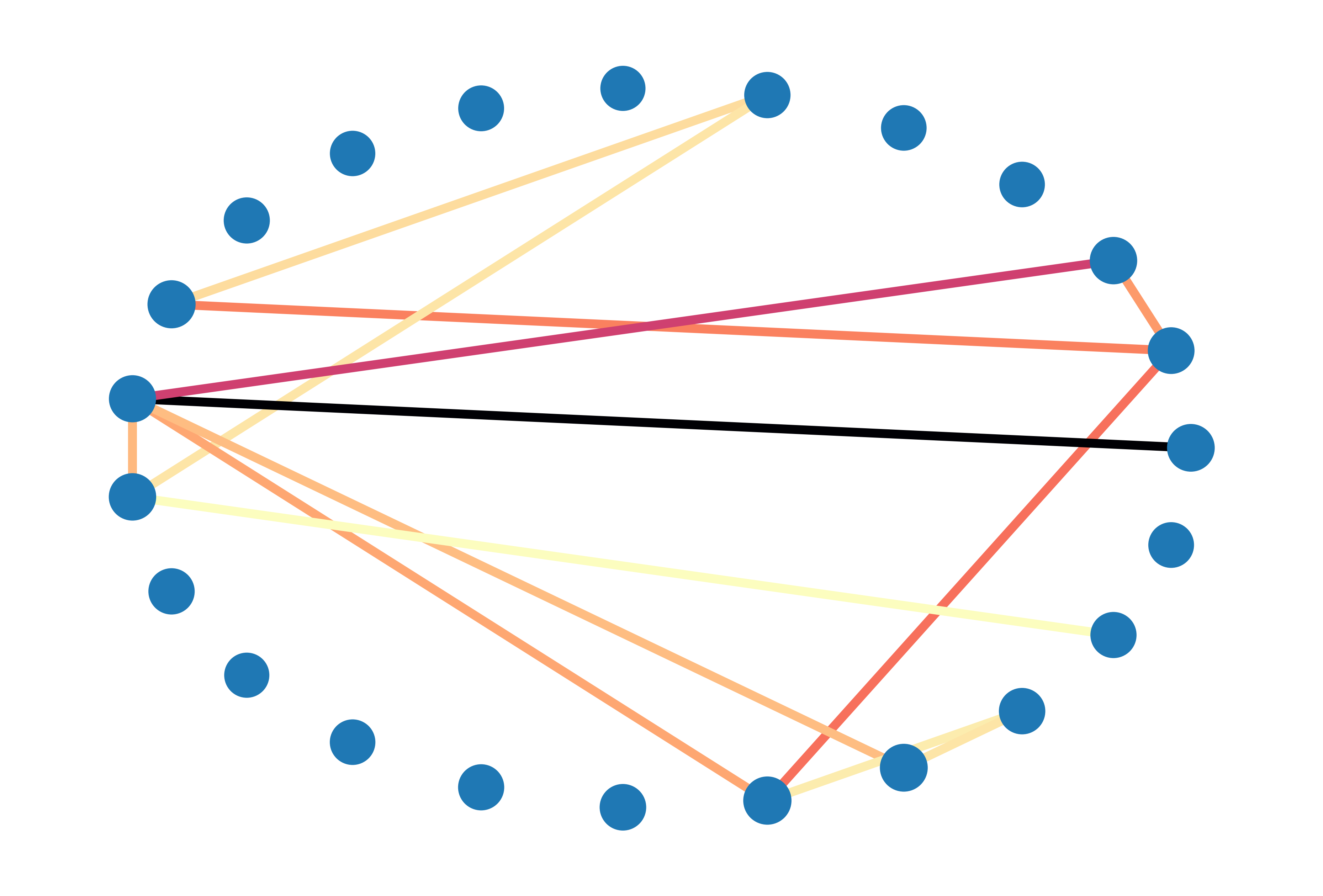}
			
			\includegraphics[width=\linewidth]{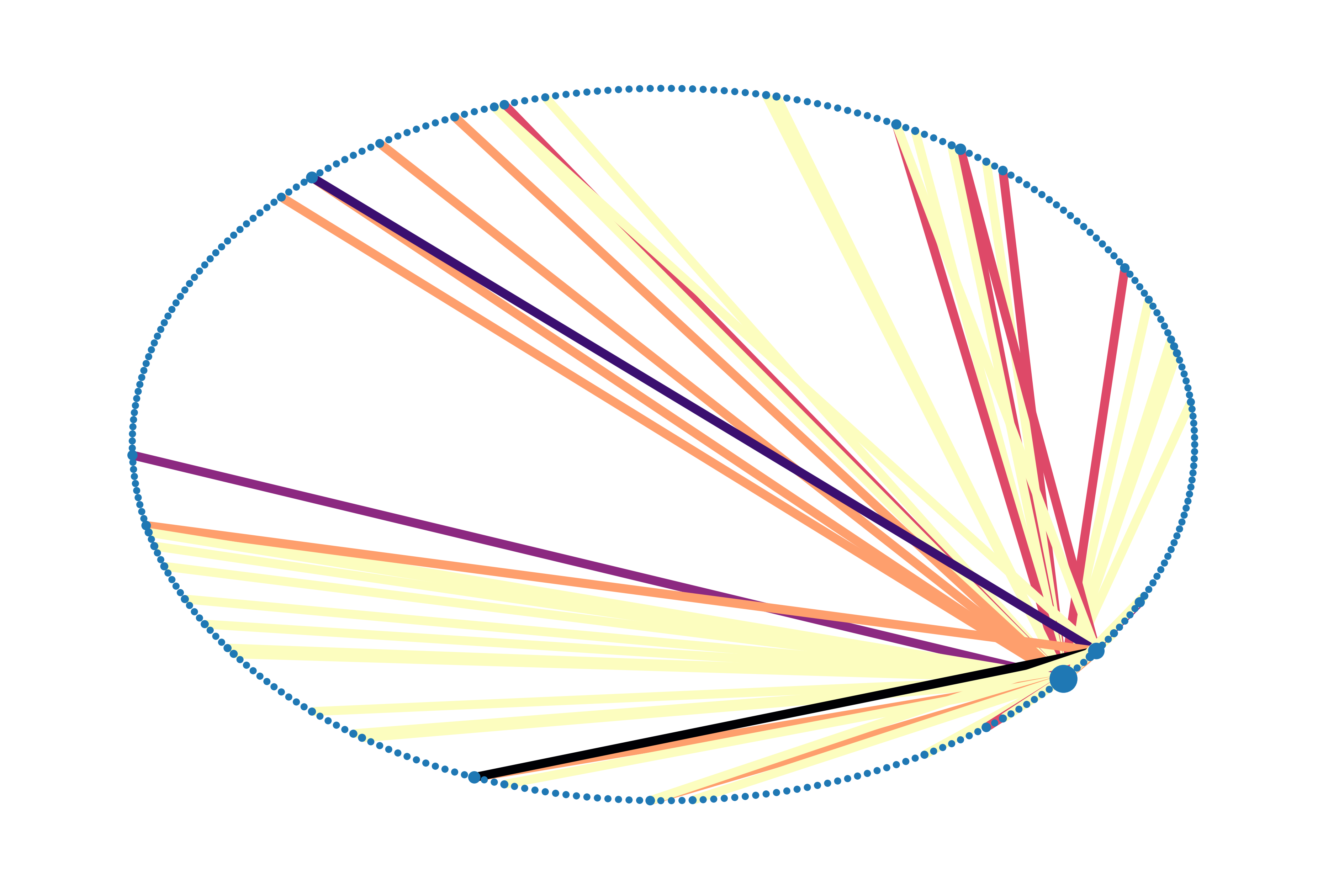}
		\end{minipage}
	} 
	\subfigure[CEP3.]{
		\begin{minipage}[b]{0.25\linewidth}
		 \includegraphics[width=\linewidth]{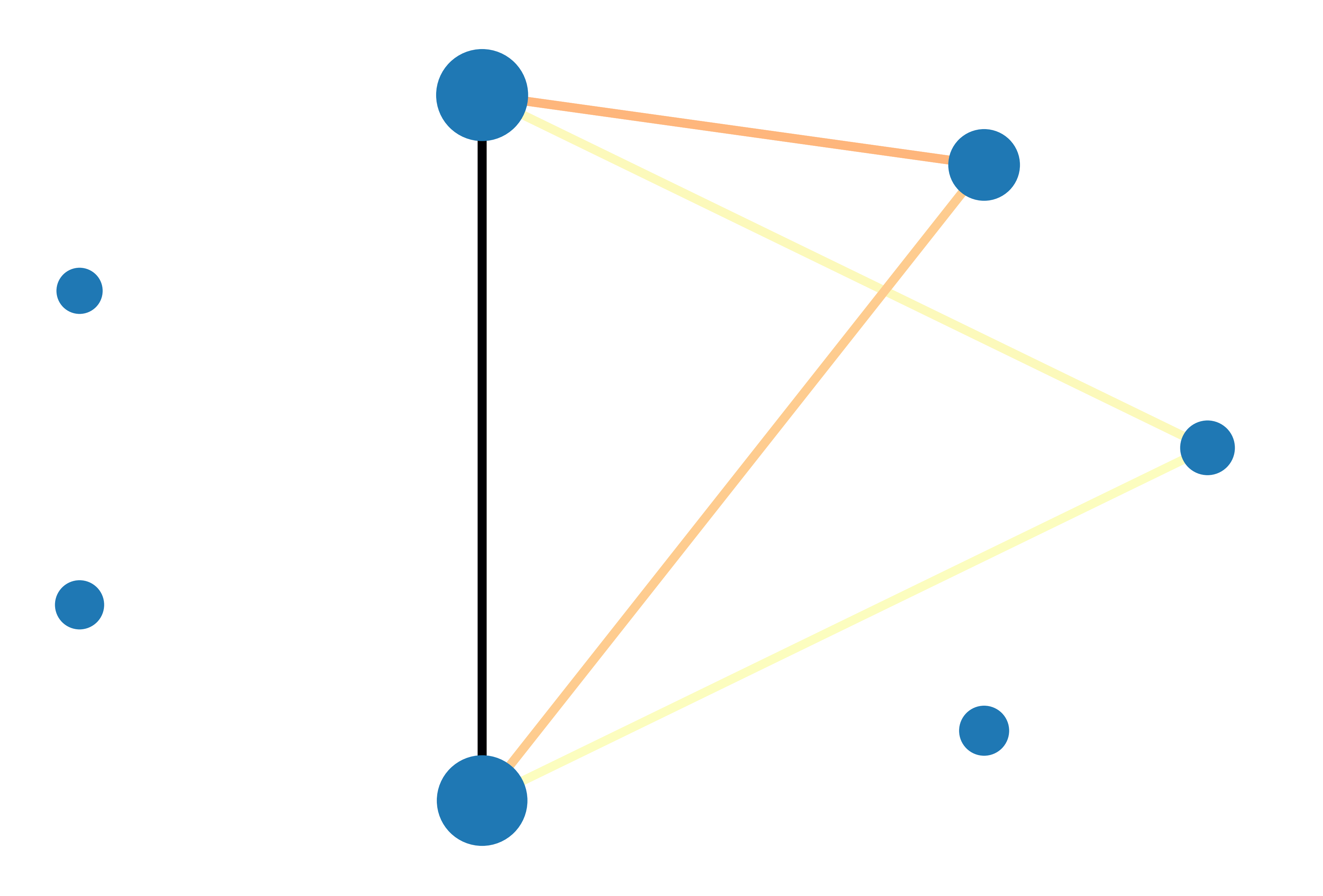}
		 
			\includegraphics[width=\linewidth]{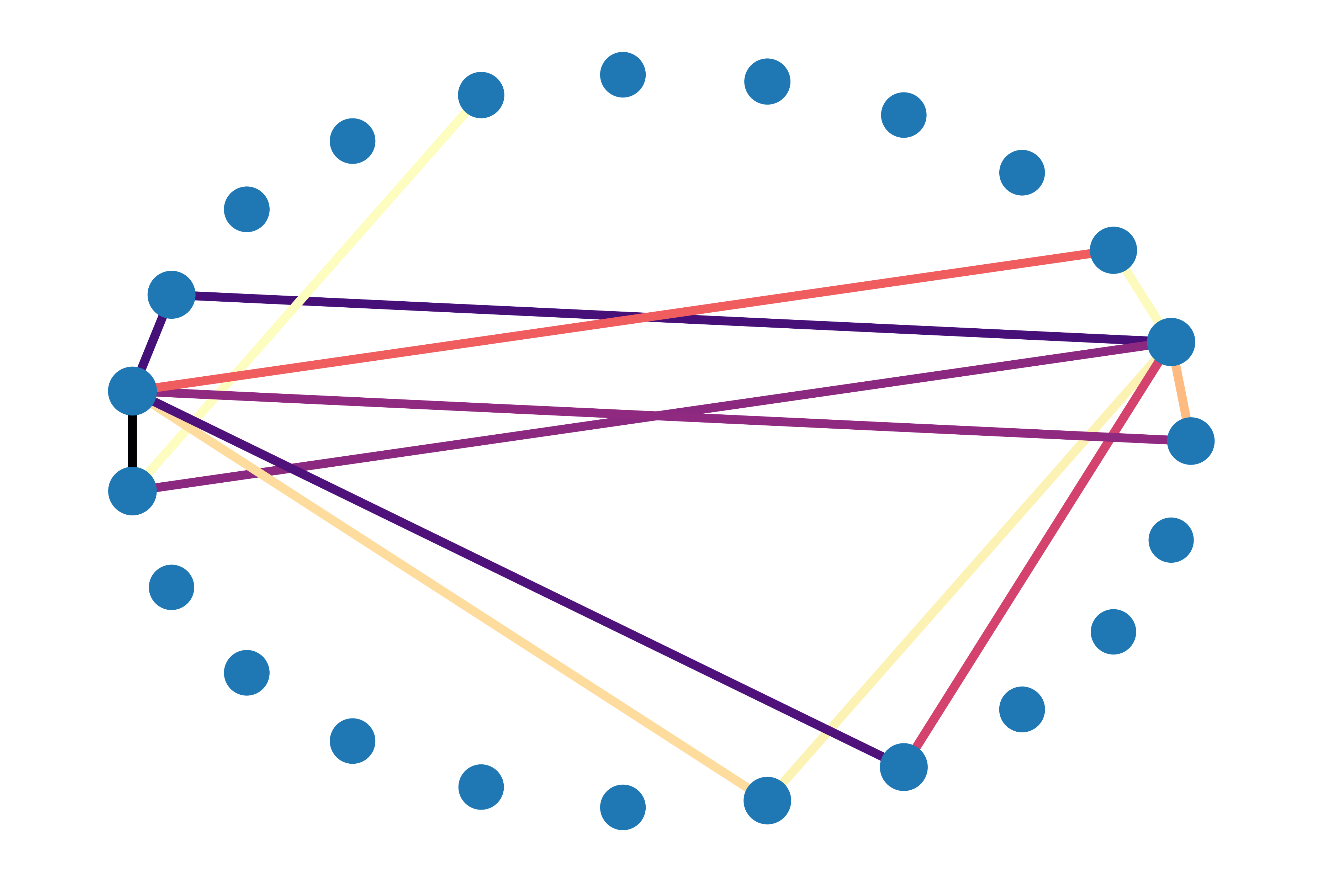}
			
			\includegraphics[width=\linewidth]{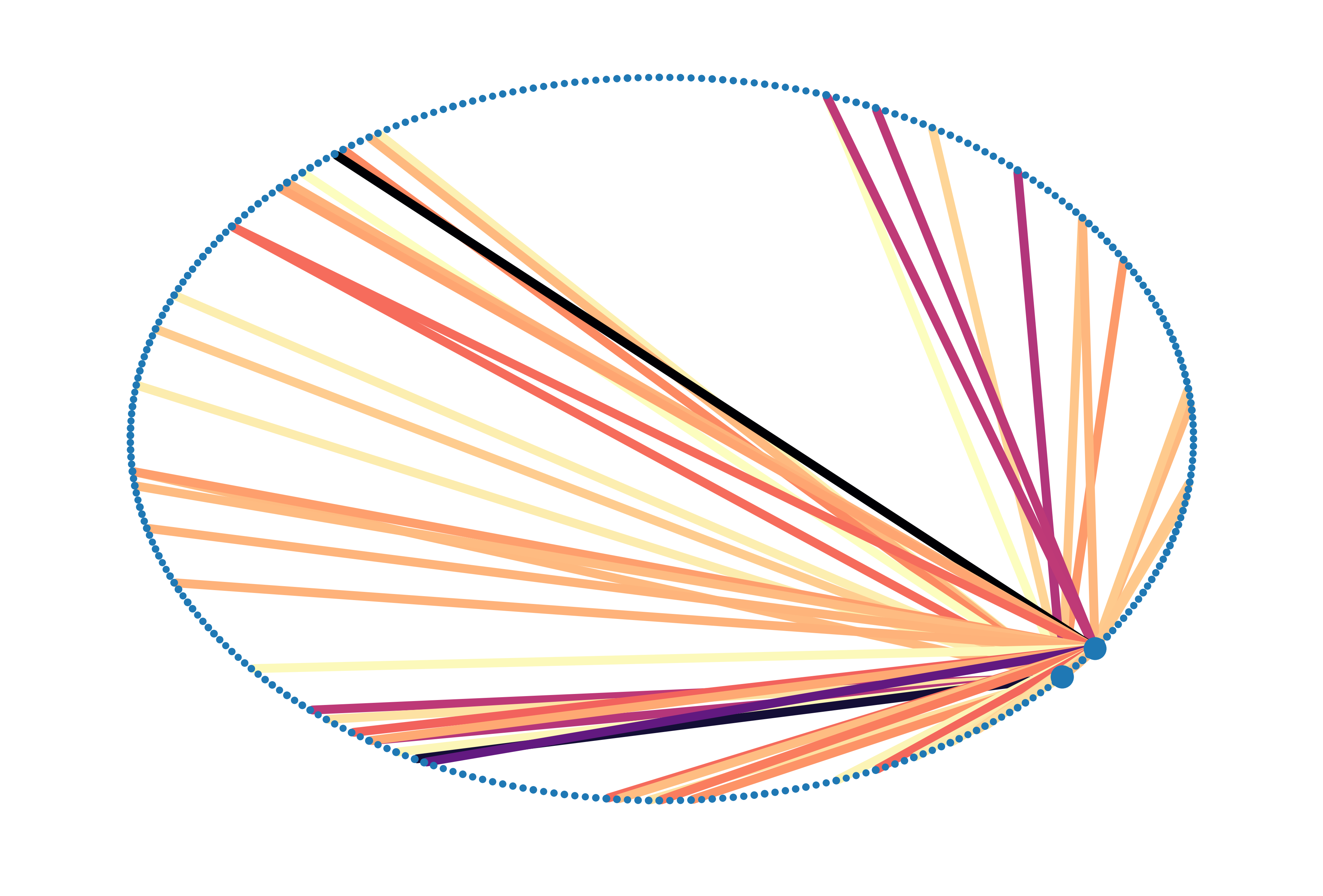}
		\end{minipage}
	}
	\subfigure[DyRep.]{
		\begin{minipage}[b]{0.25\linewidth}
			\includegraphics[width=\linewidth]{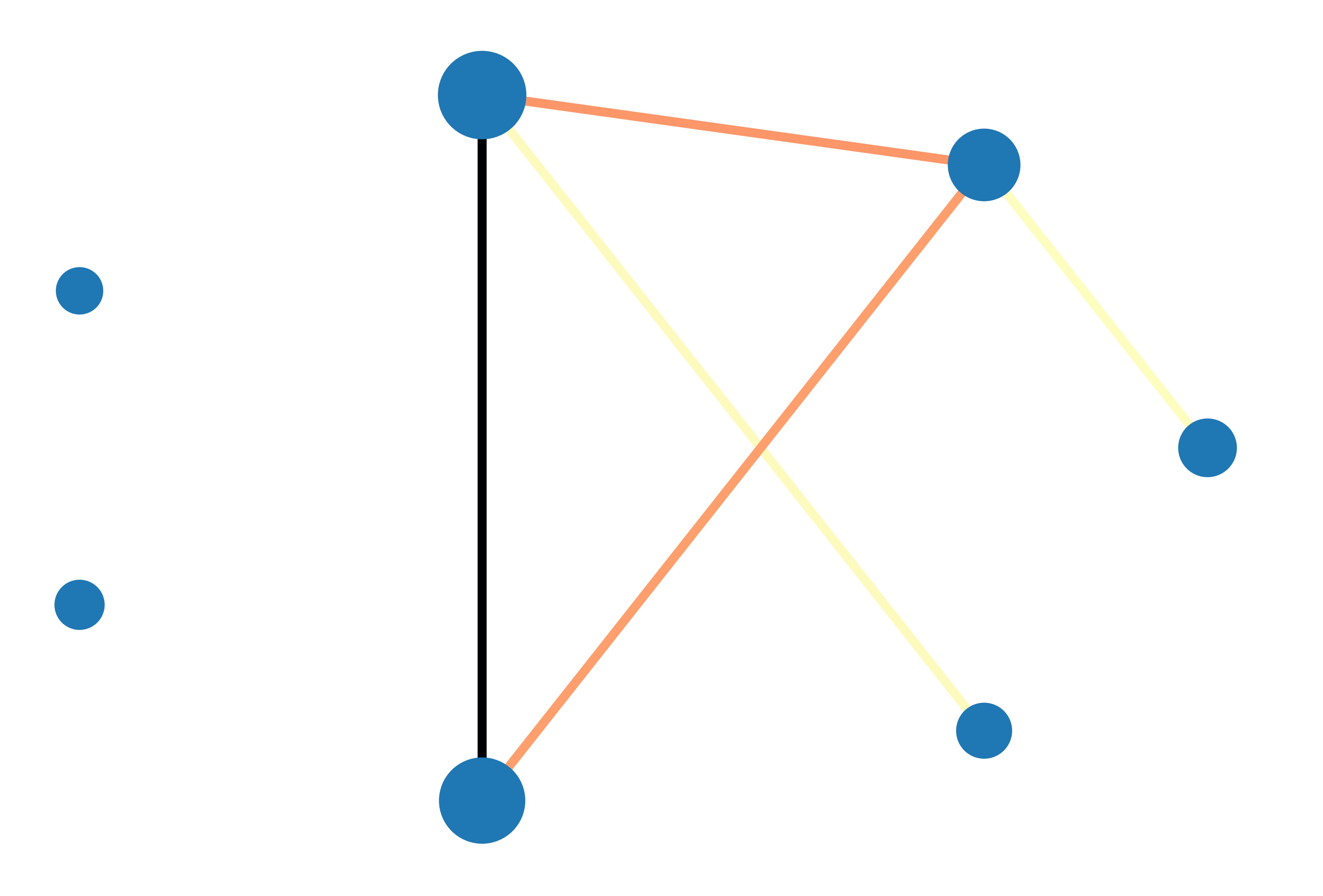}
			
			\includegraphics[width=\linewidth]{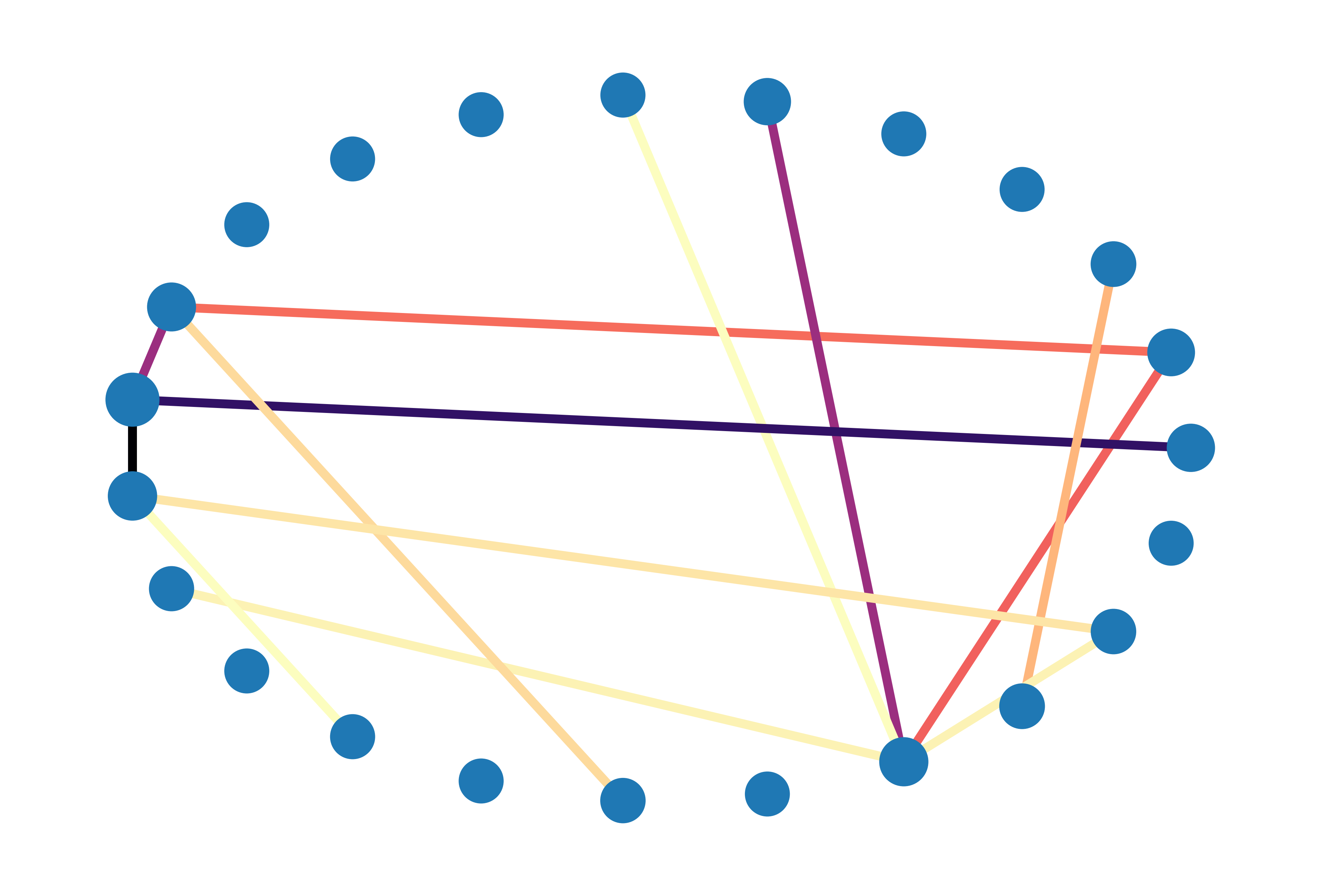}
			
			\includegraphics[width=\linewidth]{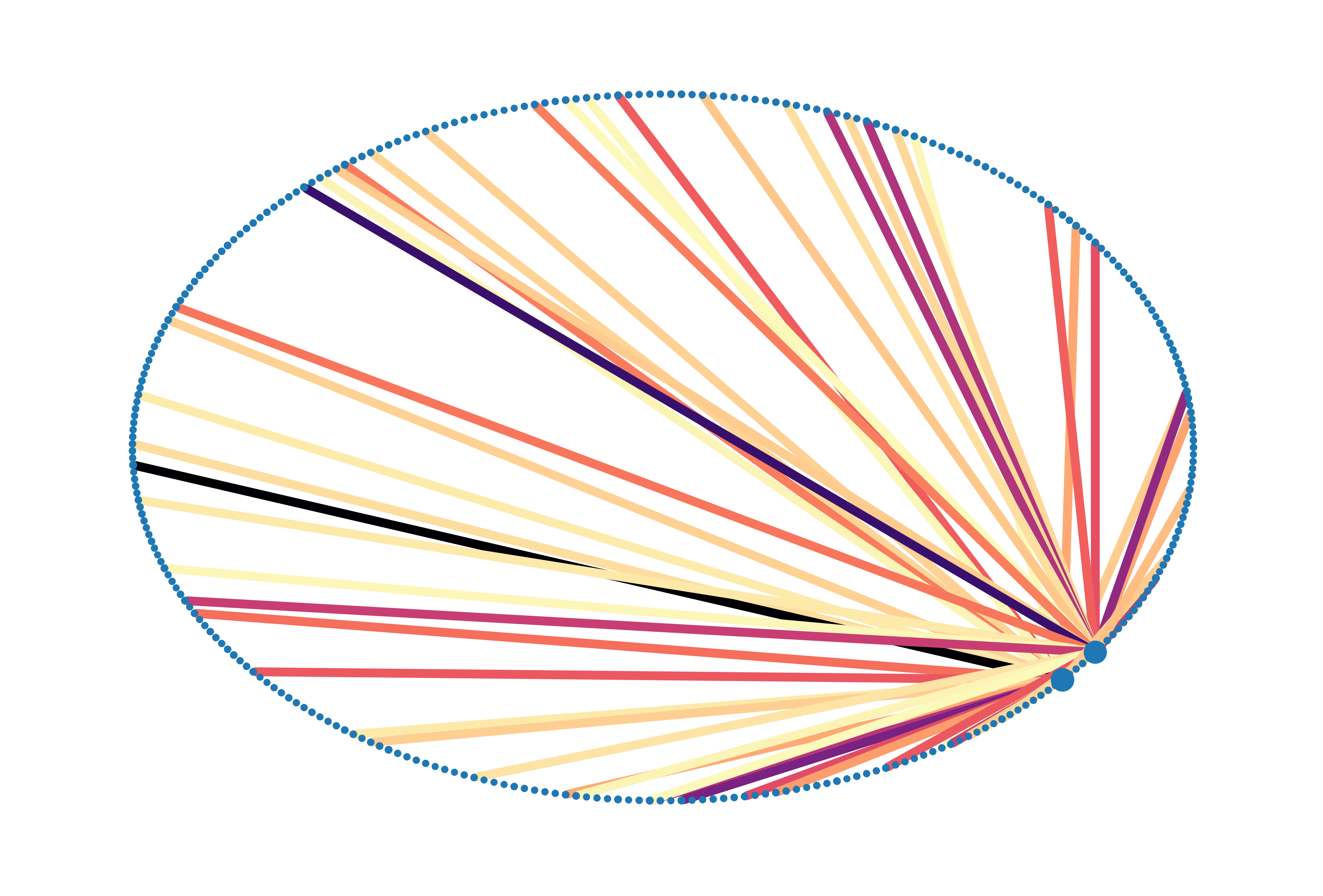}
		\end{minipage}
	}
	\subfigure{
		\begin{minipage}[b]{0.05\linewidth}
			\includegraphics[height=3.5in]{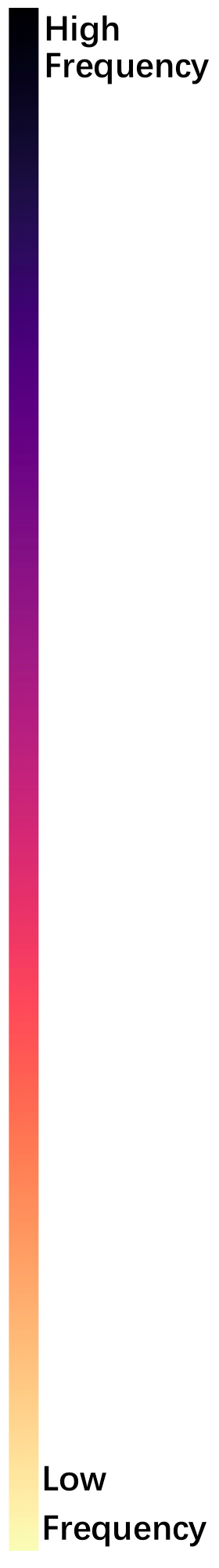}
		\end{minipage}
	}
	
	\caption{\xh{The prediction visualizations of certain communities in the whole timespan of \yh{the }test phase. The sizes of \yh{plotted }node\yh{s} indicate the\yh{ir} degrees, whereas the color\yh{s} of edges represent the connection frequenc\yh{ies}. Note that the edge colors and the node sizes can only be compared in the same row. \yh{The communities are from Github, Wikipedia and MOOC datasets, respectively, from top to bottom. } }}
	\label{fig:visualize}
\end{figure*}
\begin{table*}[!tp]
\caption{Comparison of the average and standard deviation of perplexity of incident node prediction and mean absolute error of time prediction. The smaller the MAE and Perplexity, the better the model.
Numbers are ranked by mean followed by standard deviation.  The best result is highlighted in \textbf{bold} and second best is highlighted with {underline}. \xh{The Rank column shows the average ranking in each metric and dataset\yh{ (the lower, the better)}.}}
\centering
\resizebox{\linewidth}{!}{
\begin{tabular}{@{}lrrrrrrrrr@{}}
\toprule
Datasets       & \multicolumn{2}{c}{Wikipedia}            & \multicolumn{2}{c}{Github}              & \multicolumn{2}{c}{MOOC}                    & \multicolumn{2}{c}{SocialEvo}             & \multicolumn{1}{c}{\multirow{2}{*}{Rank}} \\ \cmidrule(r){1-9}
Metric         & Perplexity          & MAE                & Perplexity         & MAE                & Perplexity            & MAE                 & Perplexity          & MAE                 & \multicolumn{1}{c}{}                      \\ \midrule
GRU+Gaussian   & 131.06 $\pm$ 11.27        & 54.54 $\pm$ 1.19         & 68.53 $\pm$ 1.18         & 59.05 $\pm$ 1.72         & 457.40 $\pm$ 6.25            & 36.49 $\pm$ 2.01          & 33.85 $\pm$ 0.27          & 131.71 $\pm$ 7.09         & 8.00                                      \\
Hawkes         & 108.00 $\pm$ 3.73            & 56.84 $\pm$ 0.31         & 74.40 $\pm$ 2.47          & 55.21 $\pm$ 0.12         & 502.31 $\pm$ 12.30           & 36.67 $\pm$ 0.29          & 45.33 $\pm$ 5.35          & 139.35 $\pm$ 0.17         & 9.50                                      \\
Poisson        & 119.19 $\pm$ 1.11         & 56.70 $\pm$ 0.11          & 61.49 $\pm$ 0.96         & 55.21 $\pm$ 0.31         & 438.61 $\pm$ 7.05           & 36.61 $\pm$ 0.78          & 40.48 $\pm$ 1.99          & 139.3 $\pm$ 1.15          & 8.25                                      \\
RMTPP w HRCHY  & 133.68 $\pm$ 2.31         & 34.15 $\pm$ 0.89         & 62.19 $\pm$ 0.88         & 55.05 $\pm$ 1.02         & 616.79 $\pm$ 25.74          & 32.29 $\pm$ 1.59          & 41.37 $\pm$ 6.55          & 140.02 $\pm$ 2.06         & 8.88                                      \\
RMTPP          & 121.67 $\pm$ 1.01         & 32.91 $\pm$ 1.90          & 67.97 $\pm$ 1.02         & 54.79 $\pm$ 0.47         & 664.07 $\pm$ 11.05          & 32.83 $\pm$ 2.40           & 37.05 $\pm$ 0.77          & 138.9 $\pm$ 2.30           & 8.00                                      \\
DyRep w AR     & 116.07 $\pm$ 4.98         & {\ul 28.74 $\pm$ 0.37}   & 54.57 $\pm$ 1.82         & {\ul 28.46 $\pm$ 0.65}   & 431.18 $\pm$ 1.18           & 29.92 $\pm$ 1.48          & 29.6 $\pm$ 1.93           & 99.96 $\pm$ 6.18          & 3.38                                      \\
DyRep          & 119.13 $\pm$ 1.02         & 30.04 $\pm$ 0.14         & 64.05 $\pm$ 0.78         & 36.97 $\pm$ 1.74         & 438.61 $\pm$ 9.28           & \textbf{13.41 $\pm$ 1.42} & 36.59 $\pm$ 3.02          & 103.01 $\pm$ 3.49         & 4.75                                      \\ \midrule
CEP3 w RNN     & {\ul 104.87 $\pm$ 8.70}    & 41.94 $\pm$ 1.89         & 60.18 $\pm$ 1.04         & 39.22 $\pm$ 2.93         & {\ul 374.77 $\pm$ 24.59}    & 20.09 $\pm$ 0.33          & {\ul 30.37 $\pm$ 4.56}    & 95.12 $\pm$ 2.25          & 3.88                                      \\
CEP3 w/o HRCHY & \textbf{98.98 $\pm$ 7.61} & \textbf{28.69 $\pm$ 0.70} & {\ul 52.04 $\pm$ 3.33}   & \textbf{26.8 $\pm$ 0.89} & \textbf{365.68 $\pm$ 28.01} & 31.87 $\pm$ 0.18          & \textbf{28.66 $\pm$ 2.74} & \textbf{79.58 $\pm$ 5.39} & \textbf{1.75}                             \\
CEP3 w/o AR    & 125.51 $\pm$ 7.64         & 39.31 $\pm$ 2.59         & 61.03 $\pm$ 1.03         & 34.03 $\pm$ 0.37         & 448.37 $\pm$ 4.34           & 21.4 $\pm$ 0.47           & 38.59 $\pm$ 1.02          & 95.21 $\pm$ 4.44          & 6.13                                      \\
CEP3           & 118.82 $\pm$ 4.30          & 32.41 $\pm$ 0.58         & \textbf{50.42 $\pm$ 0.70} & 30.93 $\pm$ 1.67         & 401.64 $\pm$ 7.06           & {\ul 17.69 $\pm$ 2.68}    & 36.8 $\pm$ 1.00             & {\ul 94.54 $\pm$ 7.31}    & {\ul 3.25}                                \\ \bottomrule
\end{tabular}
}
\label{tab:eventforecastresult}
\end{table*}

\xh{From top to bottom, Fig.~\ref{fig:visualize} visualizes the circle layout of \yh{a certain} community \yh{within the graphs of the Github, Wikipedia and MOOC datasets, respectively. }We plot the ground truth, our model's prediction and DyRep's prediction for comparison. 
\yh{The visualized graphs are generated as follows: We first apply the learned forecasting model to predict the edges using Monte Carlo sampling. This generation process is repeated for three times. We then discard generated edges that are not within the 33\% highest predicted edge probabilities and obtain the final generated graph. Generating multiple times and then discarding the less possible edges is to reduce uncertainty in each one-time generated graphs.}}

\xh{In the first row, both CEP3 and DyRep capture the triangle connection in this small community. However, the triangle is \yh{lighter} in the ground truth, \yh{which means that }DyRep over-reinforces this connection in its predictions. In the second row, the prediction result of CEP3 is more similar \yh{to the} truth, whereas DyRep generate\yh{s} a 
high-frequency purple edge \yh{which does} not exist in the original graph. In the third row, CEP3 successfully learn\yh{s} the two black edges in ground truth, but DyRep \yh{predicts} more than two links \yh{in} darker colors. }

We can see that our method successfully recognises the high-degree nodes captured many patterns of interactions as well as the evolution dynamics of the interactions graph, which includes the nodes with higher degrees. \xh{The fundamental goal of community event prediction, rather than focusing on a single local node, is to anticipate if a certain high-frequency pattern will emerge in the community \yh{from} a global \yh{perspective}. Exploring money laundering patterns~\cite{DBLP:conf/aaai/SavageWZCY17} in finance and disease transmission patterns~\cite{das2020predicting} in communities are two of the most important examples.\YHd{refs here?}}

\subsection{Ablation Study}

In Section~\ref{sec:Method} we have mentioned three variants: \textbf{CEP3 w RNN} to trade parallelized training for long-term dependency modeling with RNN based memory module, \textbf{CEP3 w/o HRCHY} to compare hierarchical (HRCHY) prediction versus joint prediction of incident nodes, and \textbf{CEP3 w/o AR} to \xh{compare using auto-regressive module versus not using.}  

\begin{figure}[tb!]
\centering
  \includegraphics[width=0.9\linewidth]{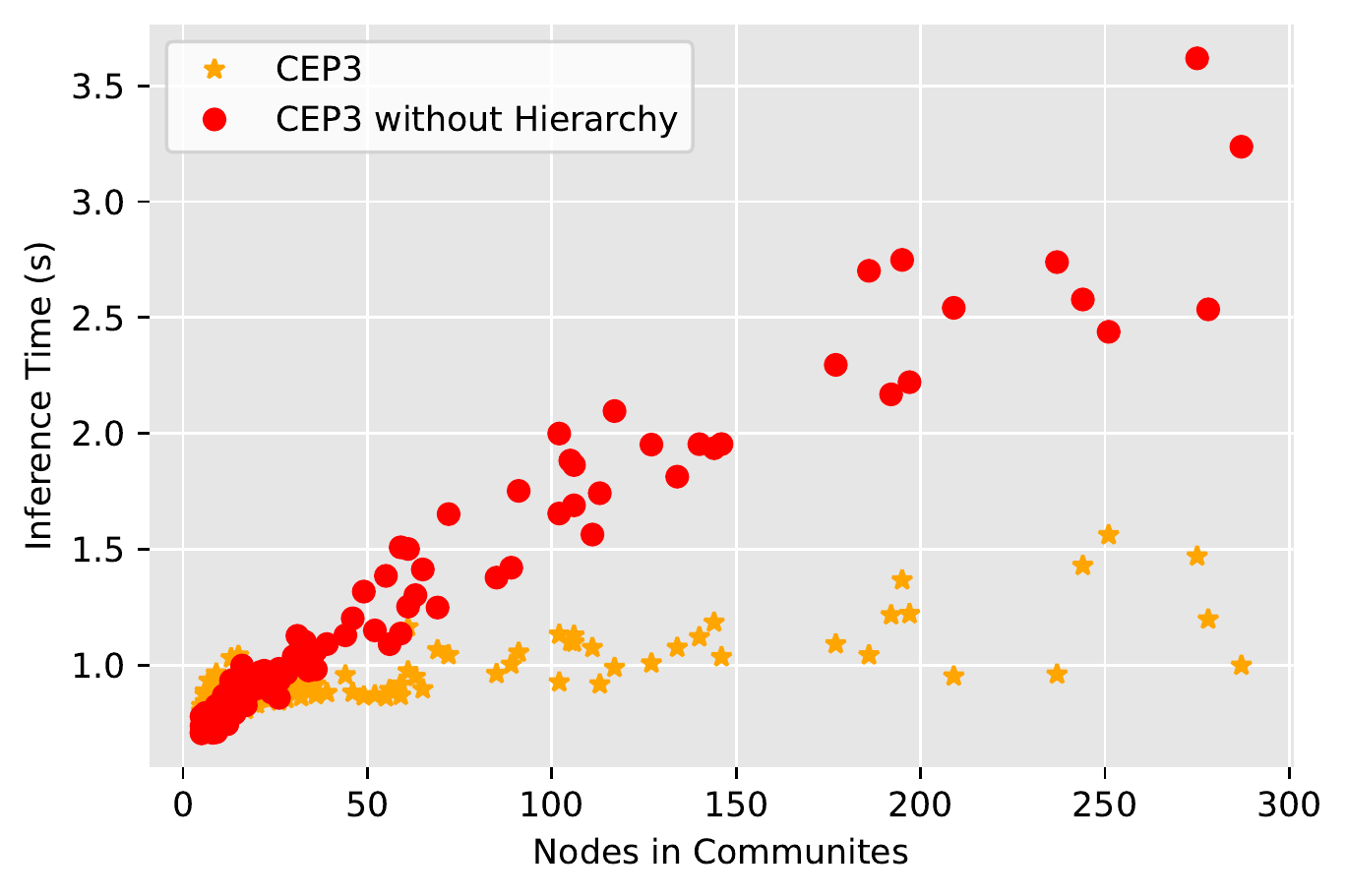}
  \caption{\xh{To demonstrate the efficiency of proposed CEP3 hierarchical probability chain, we show 1000 steps' inference time cost against the node scale. In this scatter figure, \yh{each} data point represents a community in the Wikipedia dataset.} 
  }
  \label{fig:scale}
\end{figure}

 \textbf{CEP3 w/o HRCHY.} \xh{The formulation without hierarchy structure yields a slower training and inference speed as shown in Fig.~\ref{fig:scale}. We demonstrate that CEP3 is more efficient on large communities compared to CEP3 w/o HRCHY. As is shown in Fig.~\ref{fig:scale}, computing intensity function of each possible node pair would take more time by orders of magnitude. We relax this issue by decomposing the node pair prediction problem into two node prediction sub-problems, which can be solved \yh{quicker} especially in large scale communities. 
}

\begin{figure}[tb!]
\centering
  \includegraphics[width=0.9\linewidth]{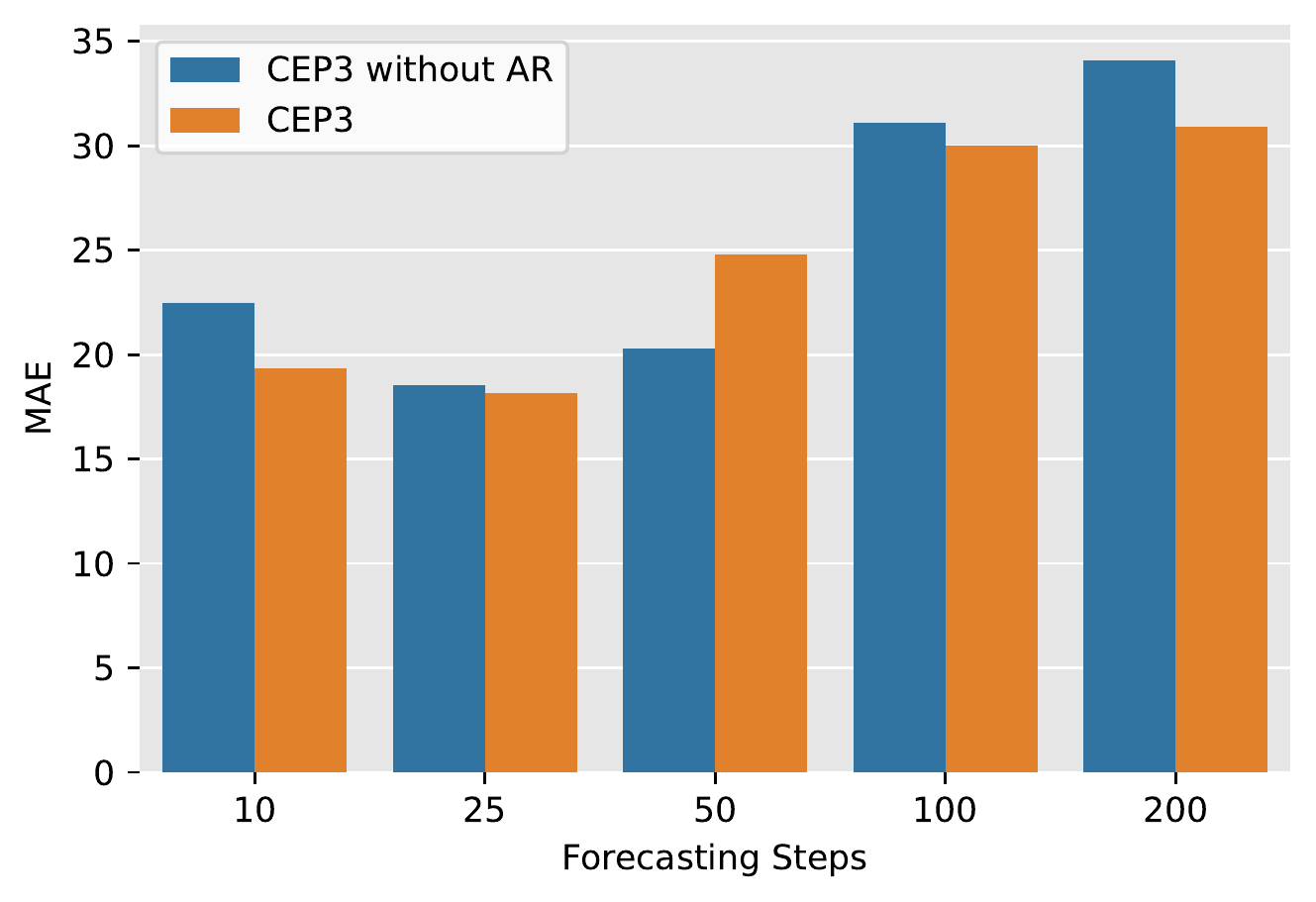}
  \caption{\xh{The \yh{number of}
  forecasting step is 
  an important parameter in almost all forecasting model\yh{s}. To further investigate the effect of\yh{ the number of} forecasting step\yh{s} and \yh{the }AR module on the CEP3 model, we run experiments with different \yh{numbers of }forecasting steps. The small MAE, the better\yh{ the} model. } 
  }
  \label{fig:steps}
\end{figure}

 \textbf{CEP3 w/o AR.} The model performance dropped significantly and yielded similar result as DyRep w AR. \xh{From Fig.~\ref{fig:steps}, we can see that the AR forecasting is not 
 \yh{helpful }in all \yh{varying numbers} of prediction steps. When the \yh{number of }``prediction steps'' is small (such as 10), CEP3 
 could just use the node embeddings at time $t_n$ to predict the events. When the \yh{number of steps} becomes larger, the systematic accumulated errors from AR gradually accumulate, leading to low MAE accuracy. And as the \yh{number} of steps increases, the initial node embedding \yh{would have little effect in} distant future events. This indicates that without AR, it may be difficult for the CEP3 model to accurately predict long-term occurrences.}
 
 \textbf{CEP3 w RNN.} \xh{The results show that the memory module brings performance improvement compared with pure CEP3, except on the Github dataset. 
 The reason is that the Github dataset is a \yh{small} dataset with a \yh{small} number of nodes and edges, and meanwhile it has a significantly longer timespan than other datasets. It means that the interaction is low-frequency in this situation, causing the insufficient memory updating. }

\subsection{Parallel Training}
\begin{figure}[tb!]
\centering
\includegraphics[width=0.9\linewidth]{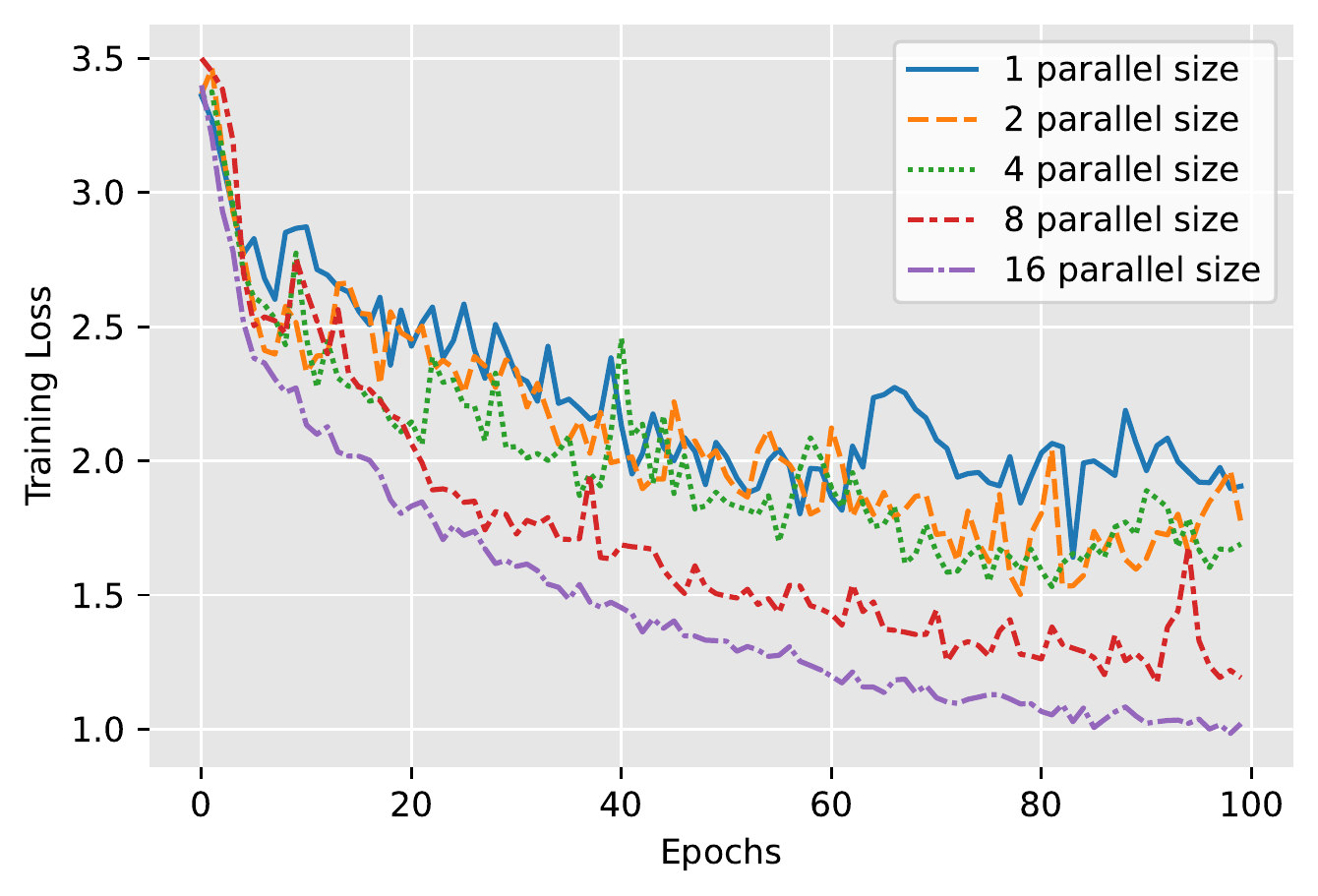}
  \caption{Training loss curve of different parallel sizes.}
  \label{fig:parallel}
\end{figure}
\xh{Inspired by the Transformer~\cite{DBLP:conf/nips/VaswaniSPUJGKP17} models in NLP, we believe it is essential to use a pure attention-based model in training temporal graph encoders.
\yh{This is because} using \yh{a }pure attention-based GNN as an encoder enables us to train multiple time windows in minibatches as described in Section~\ref{sec:Encoder}, whereas the model such as Dyrep and RMTPP cannot 
\yh{utilize }parallel training 
\yh{due to }their RNN structures.}
\xh{This property allow\yh{s} our model to 
\yh{benefit }from mini-batch training such as gradient stabilization and faster convergence. \ref{fig:parallel} shows our experiment result on\yh{ the} Wikipedia dataset with different number\yh{s} of parallel processes, \yh{which suggests }that using parallel training can increase the speed significantly without losing accuracy.}




\section{Conclusion and Future Work}
\yh{In this paper, w}e \yh{present} a novel community event forecasting task on a continuous time dynamic graph, \yh{and }set up benchmarks using adaptation of the previous work. We further propose a new model to tackle this problem utilizing graph structure\yh{s}. We also address the scalability problem when formulating\yh{ the} temporal point process on graph and reduce complexity with a hierarchical formulation. 

For future work, \yh{we are aware} that current evaluation metrics of this task \yh{focus only on either time or event type. }
\xh{Although we perform a community visualization to show the event type prediction in a time window}, a comprehensive joint metric is \yh{preferred }
to better evaluate the event forecasting task. 

\clearpage
\bibliographystyle{ACM-Reference-Format}
\bibliography{CEP3.bib}

\clearpage
\appendix


\end{document}